\documentclass{article} 
\usepackage{tencent_ailab_tech_report}
\usepackage{graphicx}
\usepackage{enumitem}
\usepackage{pgfplots}
\usepackage{tabularx}
\usepackage{caption}
\usepackage{subcaption}
\usepackage{bbm}
\usepackage{url}
\usepackage{amsmath} 
\usepackage{pifont}
\usepackage{arydshln}
\usepackage{makecell}
\usepackage{booktabs}
\usepackage{multirow} 
\usepackage{algorithm}
\usepackage{algorithmic}
\usepackage{adjustbox}
\usepackage[frozencache,cachedir=.]{minted}
\usepackage{listings}
\usepackage{xcolor}
\usepackage{lipsum}
\usepackage{wrapfig}
\usepackage{multirow}
\usepackage{amsmath, amssymb}
\usepackage{microtype}
\usepackage{hyperref}
\usepackage{url}
\usepackage{booktabs}
\definecolor{darkblue}{rgb}{0, 0, 0.5}
\hypersetup{colorlinks=true, citecolor=darkblue, linkcolor=darkblue, urlcolor=darkblue}

\DeclareUnicodeCharacter{211D}{\ensuremath{\mathbb{R}}}

\DeclareUnicodeCharacter{2264}{\ensuremath{\leq}}
\DeclareUnicodeCharacter{2265}{\ensuremath{\geq}}
\DeclareUnicodeCharacter{2192}{\ensuremath{\rightarrow}}
\DeclareUnicodeCharacter{2227}{\ensuremath{\wedge}}
\DeclareUnicodeCharacter{22A2}{\ensuremath{\vdash}}
\DeclareUnicodeCharacter{2080}{\ensuremath{_0}}
\DeclareUnicodeCharacter{2081}{\ensuremath{_1}}
\DeclareUnicodeCharacter{2082}{\ensuremath{_2}}
\DeclareUnicodeCharacter{2083}{\ensuremath{_3}}
\DeclareUnicodeCharacter{2084}{\ensuremath{_4}}
\DeclareUnicodeCharacter{2085}{\ensuremath{_5}}
\DeclareUnicodeCharacter{2086}{\ensuremath{_6}}
\DeclareUnicodeCharacter{2087}{\ensuremath{_7}}
\DeclareUnicodeCharacter{2088}{\ensuremath{_8}}
\DeclareUnicodeCharacter{2089}{\ensuremath{_9}}
\DeclareUnicodeCharacter{2200}{\ensuremath{\forall}}

\DeclareUnicodeCharacter{2090}{\ensuremath{_a}}  
\DeclareUnicodeCharacter{2091}{\ensuremath{_e}}  
\DeclareUnicodeCharacter{2092}{\ensuremath{_o}}  
\DeclareUnicodeCharacter{2093}{\ensuremath{_x}}  
\DeclareUnicodeCharacter{2211}{\ensuremath{\sum}}        
\DeclareUnicodeCharacter{2115}{\ensuremath{\mathbb{N}}}  

\title{MPS-Prover: Advancing Stepwise Theorem Proving by Multi-Perspective Search and Data Curation}

\author{
Zhenwen Liang\textsuperscript{1}\thanks{Email: \texttt{zhenwzliang@global.tencent.com}},~
Linfeng Song\textsuperscript{1},~
Yang Li\textsuperscript{2},~
Tao Yang\textsuperscript{2},~
Feng Zhang\textsuperscript{2},~
Haitao Mi\textsuperscript{1},~
Dong Yu\textsuperscript{1} \\
\\[-0.3em]
\quad \textsuperscript{1}Tencent AI Lab 
\quad \textsuperscript{2}Tencent LLM Department
}

\colmfinalcopy 
\begin{document}

\maketitle

\begin{abstract}
Automated Theorem Proving (ATP) in formal languages remains a formidable challenge in AI, demanding rigorous logical deduction and navigating vast search spaces. While large language models (LLMs) have shown promising performance, existing stepwise provers often suffer from biased search guidance, leading to inefficiencies and suboptimal proof strategies. This paper introduces the Multi-Perspective Search Prover (MPS-Prover), a novel stepwise ATP system designed to overcome these limitations. MPS-Prover incorporates two key innovations: a highly effective post-training data curation strategy that prunes approximately 40\% of redundant training data without sacrificing performance, and a multi-perspective tree search mechanism. This search integrates a learned critic model with strategically designed heuristic rules to diversify tactic selection, prevent getting trapped in unproductive states, and enhance search robustness. Extensive evaluations demonstrate that MPS-Prover achieves state-of-the-art performance on multiple challenging benchmarks, including miniF2F and ProofNet, outperforming prior 7B parameter models. Furthermore, our analyses reveal that MPS-Prover generates significantly shorter and more diverse proofs compared to existing stepwise and whole-proof methods, highlighting its efficiency and efficacy. Our work advances the capabilities of LLM-based formal reasoning and offers a robust framework and a comprehensive analysis for developing more powerful theorem provers.
\end{abstract}

\section{Introduction}

Automated Theorem Proving (ATP) is the task of automatically generating formal proofs for given mathematical or logical statements. By transforming problems into theorems in a formal language (e.g., Lean \citep{moura2021lean} or Isabelle \citep{paulson1994isabelle}) and recursively interacting with the proof assistant's engine to construct full proofs, an ATP system generates machine-verified proofs that guarantee strict logical correctness. This verifiability makes ATP indispensable for formal verification of solutions and proofs, where each reasoning step must be checked rigorously. ATP has long been viewed as a foundational and challenging problem in both AI and mathematics, as such systems can leverage massive computational power, potentially aiding mathematicians in evaluating new hypotheses and even solving open mathematical problems. The rapid progress of large language models (LLMs) has significantly advanced automated theorem proving (ATP), exemplified by AlphaProof's silver medal performance at IMO 2024 \citep{alphaproof2024ai}.

Recent research tackles these challenges by combining the reasoning abilities of LLMs with feedback from proof checkers (e.g., the Lean compiler). Two main approaches have emerged. One is \emph{whole-proof generation} \citep{wang2024theoremllama,xin2024deepseek,lin2025goedel,zhang2025leanabell,wang2025kimina}, where the LLM attempts to output an entire proof script in one shot. This leverages the model’s ability to plan with a high-level view but forgoes intermediate verification, making it prone to failures on long or intricate proofs. The second, and the focus of this paper, is \emph{stepwise formal proof generation} \citep{wu2024internlm2,li2024hunyuanprover,xin2025bfs}. Here, an LLM iteratively proposes the next proof step (a formal tactic) given the current proof state. After each step, a formal proof assistant verifies the result, ensuring the proof stays on a correct path and providing an updated proof state as feedback. This step-by-step strategy offers several advantages: it allows for continuous interaction with the proof engine, enables progressive simplification of the search target, offers higher fault tolerance (errors only require backtracking to the previous step, not restarting the entire proof), and naturally lends itself to tree search methods that explore different proof paths.

\begin{figure}[t!]
  \centering
   \includegraphics[width=0.85\textwidth]{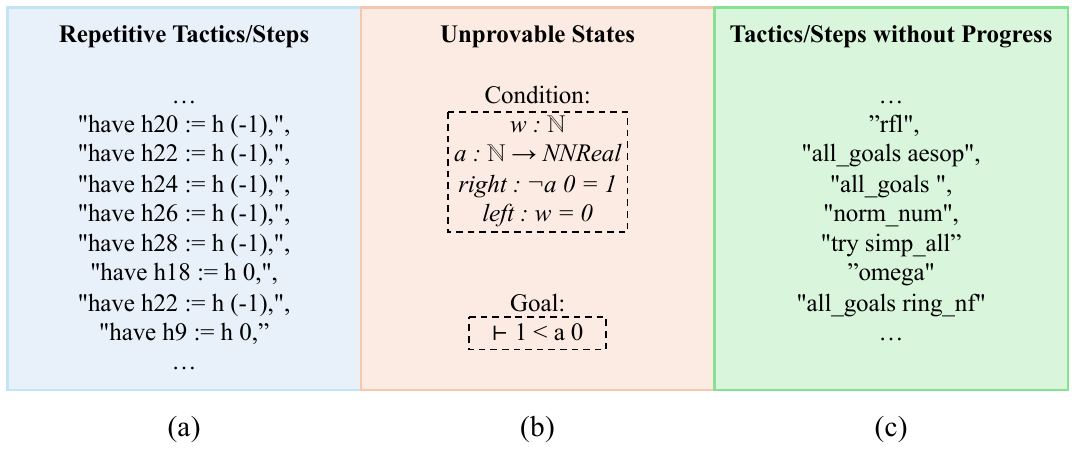}
  \caption{Common failure patterns in step-based theorem provers. (a) Repetitive steps caused by critique model over-preference for specific tactics. (b) Unprovable states resulting from incorrect tactic choices that overly simplify conditions. (c) Ineffective tactic applications that fail to make progress.}
  \vspace{-6mm}
  \label{fig:intro_revised}
\end{figure}

However, stepwise LLM-based provers face their own key challenges, as illustrated in Figure~\ref{fig:intro_revised}. First, the critic model guiding node selection in tree search can become biased. This bias often stems from the high frequency of certain "safe" or broadly applicable tactics (e.g., $have$, or general-purpose simplifiers like $aesop$ and $simp\_all$ when part of a successful sequence) in the training data. These tactics, while often part of valid proofs and less prone to immediate errors, may not always lead to the most efficient or even a correct overall proof path if the model over-relies on them, leading to stalled progress from similar tactic suggestions (Figure~\ref{fig:intro_revised}a). Second, incorrect tactic applications can lead to unprovable states by oversimplifying conditions (Figure~\ref{fig:intro_revised}b). Third, powerful but conditionally effective tactics (e.g., $aesop$, $simp\_all$) might be applied ineffectively. LLMs may propose these due to their frequent appearance in the training data or their ability to produce local simplifications that seem promising, yet they can make no progress or even obscure the path forward when the state is not genuinely suitable for such simplification (Figure~\ref{fig:intro_revised}c). While Best-First Search (BFS)-based methods \citep{li2024hunyuanprover,xin2025bfs} have shown promise in navigating this search space, their typical reliance on a single critic score for node expansion can still render them vulnerable to these failure modes, particularly the biases inherent in learned critics.

In this paper, we introduce the \emph{Multi-Perspective Search Prover (MPS-Prover)}, a novel approach designed to overcome these limitations and significantly enhance stepwise proving performance. Our first contribution is a carefully designed post-training data curation strategy. Unlike existing expert iteration approaches that uniformly add all newly proved problems, we introduce explicit rules to filter out approximately 40\% of redundant or low-value training examples, focusing the model on learning more complex reasoning patterns. This curates a higher-quality training set, leading to improved model accuracy and mitigating overfitting, especially when augmented with natural language reasoning datasets. Our second core contribution, building upon Best-First Search (BFS) methodologies \citep{li2024hunyuanprover,xin2025bfs}, is a multi-perspective tree search enhanced with strategically devised heuristic critiques. These critiques diversify tactic selection, reducing the risk of becoming trapped in repetitive, unproductive, or unprovable states by encouraging broader exploration during proof search.

Our experiments demonstrate that MPS-Prover achieves state-of-the-art performance across multiple ATP benchmarks, including miniF2F and the more challenging ProofNet. On miniF2F, MPS-Prover surpasses previous stepwise provers. Furthermore, on ProofNet, within the 7B model class, MPS-Prover outperforms all baselines, including those employing CoT reasoning. Our analyses also reveal that MPS-Prover generates significantly shorter and more diverse proofs compared to both BFS-based stepwise provers (under equivalent computational budgets) and leading whole-proof provers, highlighting the efficiency and efficacy of our multi-perspective search strategy. More specifically, the average solution length produced by MPS-Prover is only 3.44, compared to 15.91 and 52.16 for Kimina-Prover and Deepseek-Prover V2, respectively. These findings illustrate key advantages of our enhanced stepwise approach and suggest promising directions for future hybrid prover development. The primary contributions of this paper are:

1.  A novel post-training data curation strategy for stepwise provers, effectively eliminating approximately 40\% of redundant training data while achieving superior performance.

2.  The Multi-Perspective Search Prover (MPS-Prover), an innovative tree search method with heuristic critiques to enhance tactic diversity, prevent critique model bias, and improve search robustness.

3.  Demonstration of state-of-the-art performance by MPS-Prover on multiple benchmarks, including miniF2F and ProofNet, along with analyses showing generation of shorter and more diverse proofs.

\section{Method}
\subsection{Expert Iteration on Tactic Generation}
Most recent Automated Theorem Proving (ATP) systems utilize an expert iteration \citep{polu2020generative} process to collect training data, which consists of several key steps: (1) autoformalization of natural language problems into formal proofs; (2) attempting proofs or disproofs using the model trained in the previous iteration; and (3) integrating newly proved theorems and the proofing steps into subsequent training iterations.

We follow previous work to collected public available natural language problems and formal theorems and their proofs, including the Lean Workbook \citep{yinglean}, Numina \citep{li2024numinamath} and AoPS-Instruct \citep{mahdavi2025leveraging} for expert iteration. After formalizing the natural language problems, we conducted 26 iterative rounds of expert iteration. This process yielded over 30,000 proven theorems and approximately 6 million individual proving (state, step) pairs that can be used to train our prover.

\subsection{Training Data Curation}

\paragraph{Filtering Short Proofs.} 
To focus the model's learning on these more complex reasoning patterns and reduce its reliance on simple tactics, we exclude theorems from the training set that can be proven in 3 steps or fewer. The number 3 is determined by a grid search on \{2, 3, 4, 5\}.Our analysis indicated that these very short proofs predominantly rely on a limited set of elementary tactics (e.g., $rfl$, $simp\_all$, or $nlinarith$) and thus offer minimal insight into advanced problem-solving techniques. By removing these overly simplistic examples, we reduced our initial training dataset by approximately 40\%.
It is important to note that filtering these simple proofs is not expected to degrade the model's ability to solve easy problems. This is because the training data for a step-wise prover inherently includes a vast number of "late-stage" proof steps. These steps, taken when a proof is already well underway and nearing completion, often resemble the states encountered in simpler problems. Consequently, the model still receives ample exposure to simpler reasoning contexts through these intermediate steps of complex proofs.

\paragraph{Removing Ineffective Tactics.} We additionally filter out the training data where the step does not meaningfully advance the proof state. Certain tactics intended for simplification occasionally do not bring any change to the proof state, such as $aesop$, $all\_goals$, and $simp\_all$. After evaluating our dataset, we identified and removed about 5\% of such ineffective tactics. This targeted pruning encourages the model to better discern when these simplification tactics should be applied, reducing overreliance and improving proof efficiency.

\begin{figure}
  \centering
   \includegraphics[width=0.95\textwidth]{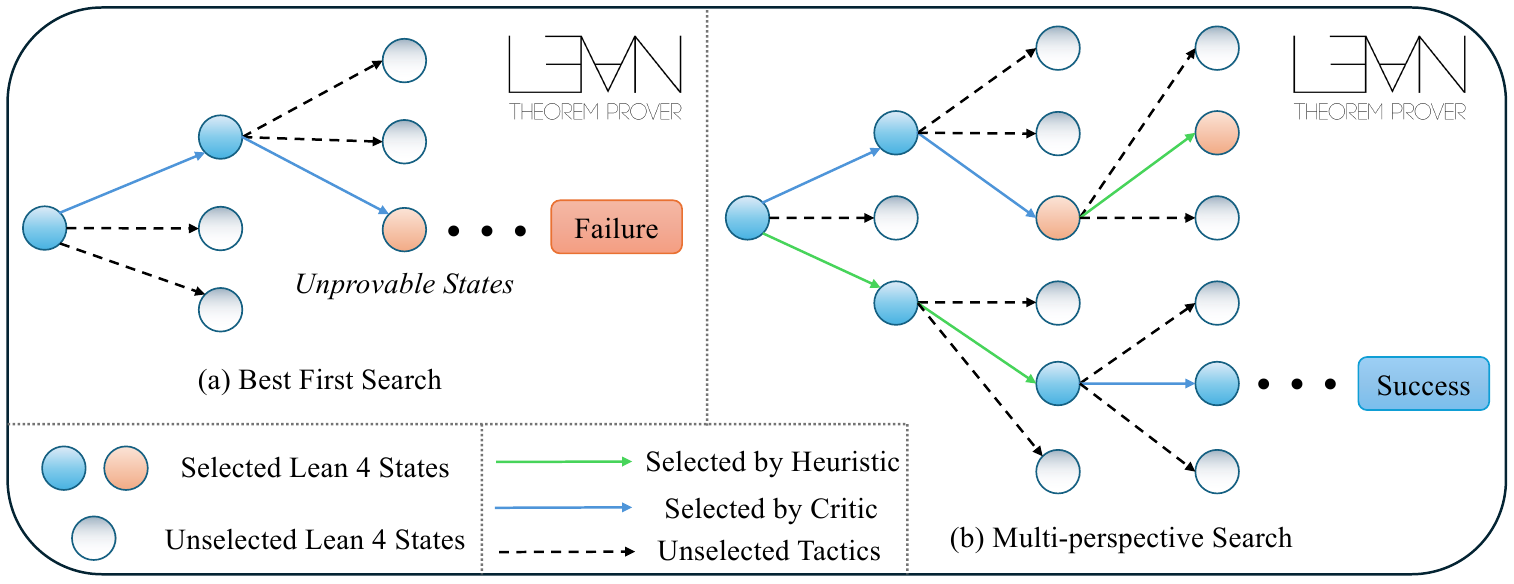}
  \caption{Search‑strategy comparison in Lean‑based proving. (a) Best‑First Search follows the single branch favoured by a learned critic; when that critic’s inherent bias selects an unprovable state or an ineffective tactic, the entire proof attempt terminates in failure. (b) Multi‑Perspective Search (MPS) evaluates each expansion step from heuristics as well as the critic, preserving a more diverse set of promising Lean 4 states and steering the prover around dead ends and toward a successful proof.}
  \vspace{-0.4cm}
  \label{fig:method}
\end{figure}

\subsection{Multi-Perspective Tree Search}
As illustrated in Figure \ref{fig:method}, the traditional BFS approach selects nodes based solely on the best critic scores. Following \cite{li2024hunyuanprover}, our critic model is trained using a hierarchical, tree-based distance prediction method to enhance its guidance capabilities during proof searches. The output from the critic model guides the proof search tree by indicating proximity to the completion of the proof—the smaller the predicted distance, the closer the node is to a successful proof.

While the critic model significantly enhances decision-making during most search steps, it can sometimes fail, resulting in unprovable states and reduced search efficiency. For instance, the proof-by-contradiction tactic, although powerful, can substantially alter the proof goal and make the theorem unprovable if misapplied. Additionally, sometimes critic models tend to frequently propose similar tactics, creating repetitive and ineffective local minima. As illustrated in Figure~\ref{fig:method}(a), the proof search is guided by the critic model to reach an intermediate node where the state becomes unprovable, leading to wasted effort on subsequent steps and ultimately resulting in a failed proof attempt. To enhance the diversity of the guiding signals, we introduce three heuristic selection rules:

\paragraph{Tactic Effectiveness Scoring.} We assign different scores to tactics based on their perceived efficacy in advancing the proof. Generally, tactics that introduce new, substantive assumptions or significantly restructure the proof goal, such as $rcases$, $intro$, $contrapose$, $induction$, or proof by $contradiction$ (when appropriately applied), are assigned higher scores. These are often tactics that can unlock new reasoning pathways or simplify the problem by breaking it down. Conversely, auxiliary tactics or those focused on more localized simplifications, like $norm\_num$ and $simp\_all$, receive lower scores. While tactics like proof by contradiction can be problematic if misapplied by the critic model alone (as noted earlier), its inclusion with a high score in this heuristic perspective ensures it remains a viable option for exploration when potentially beneficial. These scores are manually assigned based on human expert experience in Lean theorem proving. A detailed table of these tactic scores is provided in Appendix~\ref{app:score}. 

\paragraph{Minimizing Case Splits.} We select tactics that result in the fewest occurrences of \texttt{case} within the state string. While tactics like \texttt{induction}, \texttt{constructor}, and \texttt{split} are beneficial under specific circumstances, excessive case splitting complicates proof states. This heuristic encourages simpler, more manageable proof states.

\paragraph{Shortest State Preference.} We prioritize tactics leading to shorter Lean 4 state strings. Similar to minimizing cases, this heuristic favors simpler, more straightforward states, facilitating more efficient proof completion.

As shown in Figure \ref{fig:method}, our tree search maintains up to four nodes for each expansion step. Specifically, from the set of nodes selected in the previous iteration, we generate $N_{samples}$ candidate tactics for each using the LLM. This results in a larger pool of potential next states (e.g., if 4 nodes were selected and $N_{samples}=8$, we'd have $4 \times 8 = 32$ candidate next states). From this expanded pool: 1.  One node is selected based on the critic model's prediction (i.e., the one with the smallest predicted distance to completion). 2.  Three additional nodes are selected based on our heuristic rules. Each heuristic rule evaluates all candidate next states and picks the one that best satisfies its criterion. If different perspectives select the same node, we only retain it once, meaning that in such cases, fewer than four unique nodes might be carried forward to the next search iteration. 

We acknowledge that each heuristic rule, including the critic model, has inherent biases and limitations, favoring certain proof tactics or states. However, by concurrently applying these criteria, we substantially enhance the diversity of each search layer, ensuring promising nodes are retained rather than overlooked due to single-criterion bias. 
\vspace{-0.3em}
\section{Experiment}
\vspace{-0.3em}
\paragraph{Benchmarks}
To comprehensively evaluate our prover, we utilize two widely recognized benchmarks. 
1. miniF2F \citep{zheng2022miniF2F}: This is the standard benchmark in the ATP community. The problems are sourced from mathematics competitions (AMC, AIME, IMO) as well as high-school and undergraduate curricula. We use the latest version available from the Huggingface Numina repository\footnote{\url{https://huggingface.co/datasets/AI-MO/miniF2F_test}}, which corrects eight errors identified in the original dataset.
2. ProofNet \citep{azerbayev2023proofnet}: This benchmark consists of 371 problems, characteristic of undergraduate-level mathematics. We report performance on its test split.

\vspace{-0.25em}
\paragraph{Supervised Fine-tuning}
We employ supervised fine-tuning (SFT) on Qwen2.5-Math-7B-base. The SFT dataset is a composition of: (1) step-by-step proof data generated during expert iteration and curated by our proposed filtering techniques; (2) whole proof data, formed by concatenating the accepted proof steps; and (3) data for training the distance critic model for our search algorithm. This aggregated dataset amounts to approximately 3.5 million question-answer pairs after applying our training data filtering method. The model was trained for 3 epochs using a cosine learning rate scheduler with a maximum learning rate of $2 \times 10^{-5}$. We utilized a cumulative batch size of 256. The training was performed on 8 * H20 80G GPUs, with a total training duration of about 3 days.

\vspace{-0.25em}
\paragraph{Evaluation Setup}
All evaluations are conducted using Lean version 4.16.0. For interaction between the LLM and the Lean proof assistant, we utilize the \emph{repl} tool\footnote{\url{https://github.com/leanprover-community/repl}}, which facilitates step-by-step execution. To ensure the rigorous correctness of generated proofs, especially since repl might occasionally misinterpret certain erroneous steps as valid during step-wise execution, we perform a final verification. This involves concatenating all generated steps to form a complete proof script, which is then checked by the Lean compiler.
The timeout for executing a whole search is set to 3600 seconds, while the per-step tactic execution timeout is 60 seconds. Our search budget per problem is defined by the total number of tactic candidates explored. This is given by $N_{pass} \times N_{perspectives} \times N_{max\_iter} \times N_{samples}$, where $N_{pass}$ is the number of independent search trials for a problem (e.g., for pass@k, $N_{pass}=k$), $N_{perspectives}=4$, $N_{max\_iter}=800$ (maximum search iteration), and $N_{samples}=8$ (LLM samples per selected node). The "accumulative" search strategy, following \citet{xin2025bfs} (BFS-Prover), denotes an incremental evaluation protocol to assess the model's maximum potential. In this setup, we keep searching and each search iteration focuses exclusively on problems that were not solved in prior iterations. 

\vspace{-0.25em}
\subsection{Main Results}
\vspace{-0.25em}

We conduct extensive experiments to evaluate MPS-Prover against state-of-the-art methods on standard benchmarks. Our primary results on miniF2F are summarized in Table \ref{tab:wholeproof-step}, with baseline details in Appendix \ref{app:baseline}. Our method achieves the best performance among all step-level solvers evaluated. Specifically on miniF2F, MPS-Prover successfully proves 185 out of 244 problems (75.82\% accuracy), demonstrating a significant improvement over the previous state-of-the-art step-prover, BFS-prover.

\begin{table}[t]
\small
\centering
\caption{Comparison of Whole-Proof and Step-level Provers on miniF2F-test.}
\vspace{0.2cm}
\begin{adjustbox}{max width=\textwidth}
\begin{tabular}{lccc}
\toprule
\textbf{Method} & \textbf{Model Size} & \textbf{Sample Budget} & \textbf{Accuracy} \\
\midrule
\multicolumn{4}{l}{\textit{Large Whole-Proof Provers}} \\
\midrule
Kimina-Prover-Preview \citep{wang2025kimina} & 72B & $8192$ & 80.74\% \\
DeepSeek-Prover-V2 (non-CoT) \citep{ren2025deepseek} & 671B & $8192$ & 78.30\% \\
DeepSeek-Prover-V2 (CoT) \citep{ren2025deepseek} & 671B & $8192$ & \textbf{88.90\%} \\
\midrule
\multicolumn{4}{l}{\textit{Small Whole-Proof Provers}} \\
\midrule
Leanabell-Prover-GD-RL \citep{zhang2025leanabell} & 7B & $128$ & 61.1\% \\
Goedel-Prover-SFT \citep{lin2025goedel} & 7B & $25600$ & 64.7\% \\
STP \citep{dong2025stp} & 7B & 25600 & 67.6\% \\
Kimina-Prover-Preview-Distill \citep{wang2025kimina} & 7B & $1024$ & 70.8\% \\
DeepSeek-Prover-V2 (Distilled, non-CoT) \citep{ren2025deepseek} & 7B & $8192$ & 75.0\% \\
DeepSeek-Prover-V2 (Distilled, CoT) \citep{ren2025deepseek} & 7B & $8192$ & 82.0\% \\
\midrule
\multicolumn{4}{l}{\textit{Step-level Provers}} \\
\midrule
InternLM2.5-StepProver + BFS + CG \citep{wu2024internlm2} & 7B & $256 \times 32 \times 600$ & 65.9\% \\
HunyuanProver + BFS + DC \citep{li2024hunyuanprover} & 7B & $600 \times 8 \times 400$ & 68.4\% \\
BFS-Prover \citep{xin2025bfs} & 7B & $2048 \times 2 \times 600$ & 70.83\% \\
BFS-Prover \citep{xin2025bfs} & 7B & Accumulative & 72.54\% \\
\midrule
\multirow{5}{*}{MPS-Prover (Ours)} & \multirow{5}{*}{7B} & $1 \times 4 \times 800 \times 8$ & 67.62\% \\
 &  & $4 \times 4 \times 800 \times 8$ & 68.44\% \\
 &  & $16 \times 4 \times 800 \times 8$ & 70.08\% \\
 &  & $64 \times 4 \times 800 \times 8$ & 72.54\% \\
 &  & Accumulative & \textbf{75.82\%} \\
\bottomrule
\vspace{-0.9cm}
\end{tabular}
\end{adjustbox}
\label{tab:wholeproof-step}
\end{table}

When considering all models within the 7B parameter class (both whole-proof and step-wise), our model's performance is only surpassed by DeepSeek-Prover-V2 (Distilled, CoT). We posit this is expected, as their 7B model is distilled from a significantly larger model, a process known to often yield performance exceeding that of models trained natively at the smaller scale \citep{guo2025deepseek}. In contrast, our model is trained directly via iterative refinement at the 7B scale. This comparison highlights the strong performance achieved by our method and suggests substantial potential for further improvement by leveraging larger base models or incorporating techniques like Chain-of-Thought (CoT) reasoning during tactic generation.

\begin{wraptable}{r}{0.6\textwidth} 
    \centering
    \vspace{-1.2em} 
    \footnotesize 
    \caption{ProofNet-test performance of different 7B models (max budget).}
    \label{tab:proofnet_performance}
    \begin{adjustbox}{max width=0.6\textwidth}
    \begin{tabular}{lc} 
    \toprule
    \textbf{Method (7B models)} & \textbf{Performance} \\ 
    \midrule
    Goedel-Prover-SFT \citep{lin2025goedel} & 15.6\% \\
    STP \citep{dong2025stp}& 26.9\% \\
    Deepseek-Prover-V1.5-RL \citep{xin2024deepseek} & 25.3\% \\
    DeepSeek-Prover-V2 (non-CoT) \citep{ren2025deepseek} & 24.7\% \\
    DeepSeek-Prover-V2 (CoT) \citep{ren2025deepseek} & 29.6\% \\
    \midrule
    \textbf{MPS-Prover (Ours)} & \textbf{32.97\%} \\
    \bottomrule
    \vspace{-0.7cm}
    \end{tabular}
    \end{adjustbox}
\end{wraptable}

Another noteworthy finding is the strong performance of MPS-Prover even under constrained search budgets. At the minimum search budget evaluated, our model achieves a pass rate of 67.62\% on miniF2F, significantly outperforming InternLM (50.7\%) and Hunyuan Prover (59.84\%) under similar minimal conditions. Impressively, this base performance already exceeds the maximum reported performance of several strong baselines, such as Goedel-prover and InternLM2.5-StepProver. This indicates that our approach exhibits excellent stability and efficiency, capable of achieving competitive results without necessitating exhaustive search efforts.

We also evaluate the performance of MPS-Prover on the ProofNet benchmark, which is generally considered more challenging than miniF2F, featuring undergraduate-level mathematics problems that demand more intricate reasoning. Table~\ref{tab:proofnet_performance} presents a comparison of our method against other state-of-the-art 7B parameter models that have reported results on the ProofNet-test split. For a fair comparison, all models, including ours, were evaluated using their respective maximum reported sampling budgets. As can be observed, MPS-Prover achieves a success rate of 32.97\%, surpassing all other 7B baseline models. Notably, our approach outperforms even DeepSeek-Prover-V2 with Chain-of-Thought (CoT) reasoning.

\vspace{-0.25em}
\subsection{Comparison under Fixed Budgets}
\vspace{-0.25em}
A crucial aspect of evaluating search algorithms is their performance relative to computational resources. Our Multi-Perspective Search (MPS) inherently explores more branches per iteration compared to standard Best-First Search (BFS) with tree-based distance prediction as critic \citep{li2024hunyuanprover}. Specifically, MPS expands four nodes (one from the critic model and three from heuristic rules) for each selected state in a pass, whereas BFS typically expands only the single best node according to its criterion. Therefore, to ensure a fair comparison under approximately equivalent computational budgets, we compare the performance of MPS at pass@k against BFS at pass@4k.

As shown in Figure \ref{fig:mps_vs_bfs_independent}, MPS consistently outperforms BFS when allocated similar computational resources. At the lowest budget (MPS pass@1 vs. BFS pass@4), MPS achieves a success rate of 67.62\% (165/244), slightly edging out BFS's 66.39\% (162/244). This advantage becomes more pronounced as the budget increases. This consistent gap highlights the effectiveness of the diverse exploration strategy employed by MPS. By considering multiple perspectives (critic score + heuristics) at each step, MPS is less prone to getting stuck in local optima compared to the single-criterion approach of BFS, leading to a higher accuracy within a given computational budget.

\vspace{-0.25em}
\subsection{Ablation Study}
\begin{figure}[htb]
    \centering
    \begin{minipage}[t]{0.50\textwidth}
        \centering
        \includegraphics[width=\linewidth]{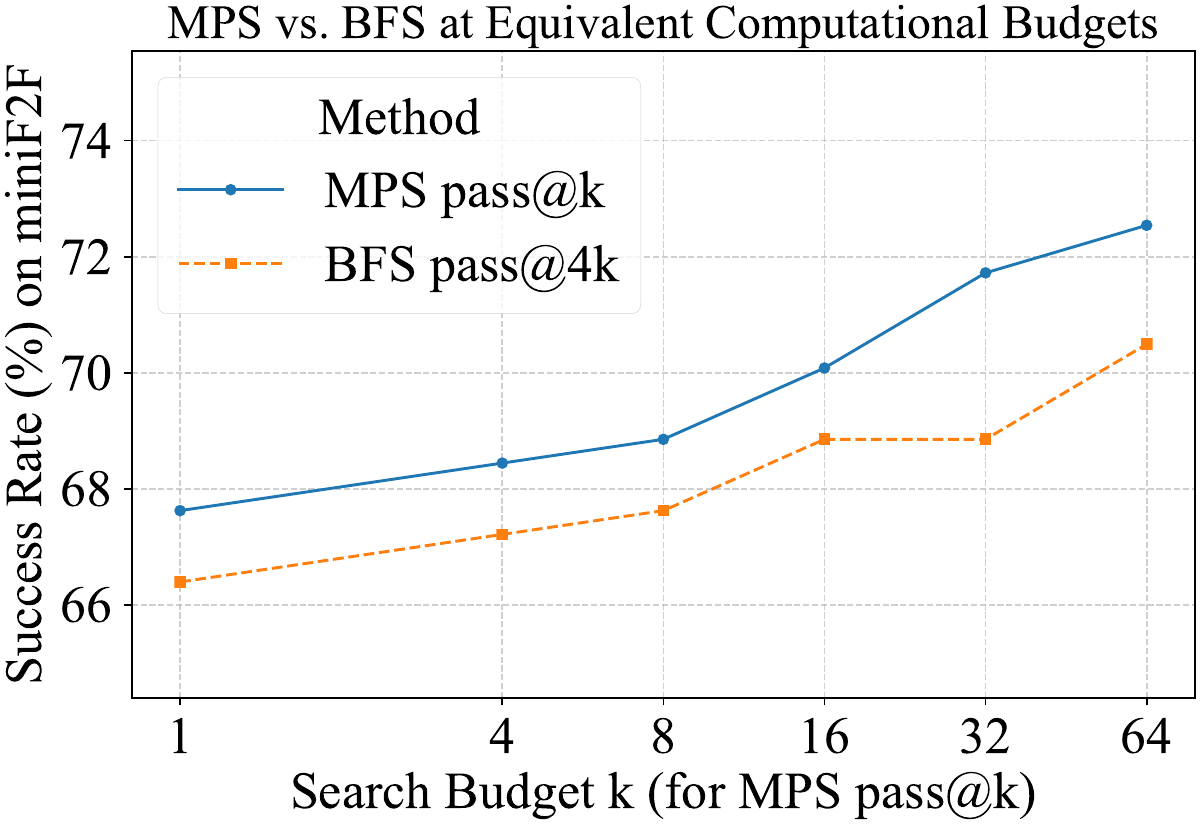}
        \caption{Performance comparison under equivalent computational budgets.}
        \vspace{-42mm}
        \label{fig:mps_vs_bfs_independent}
    \end{minipage}
    \hfill
    \begin{minipage}[t]{0.45\textwidth}
        \centering
        \vspace{-52mm} 
        \captionof{table}{Ablation study on miniF2F. Performance is problems proved (out of 244) under a fixed budget (MPS pass@64 equivalent). "w/o" indicates removing the specified component or replacing its guidance with random selection for heuristics. }
        \label{tab:ablation_study_independent}
        \resizebox{\linewidth}{!}{%
            \begin{tabular}{lc}
            \toprule
            Method Variation & Proved (n/244) \\
            \midrule
            MPS-Prover (Full Method) & 177/244 \\
            \midrule
            \quad w/o Data Curation & 176/244 \\
            \midrule
            \quad w/o Critic Model & 164/244 \\
            \quad w/o Tactic Eff. & 174/244 \\
            \quad w/o Min. Cases & 173/244 \\
            \quad w/o Short. State & 172/244 \\ 
            \bottomrule
            \end{tabular}
        }
    \end{minipage}
\end{figure}

\vspace{-0.25em}
To understand the contribution of each component in our proposed MPS-Prover, we conduct a comprehensive ablation study. We evaluate the performance of our system by systematically removing or altering key components while keeping the total computational budget fixed. This budget is equivalent to our full method's pass@64 setting (i.e., $64 \times 4 \times 800 \times 8$). The experiments are performed on the miniF2F benchmark, and the results are presented in Table \ref{tab:ablation_study_independent}.

The ablation study reveals several key insights. First, removing our post-training data curation strategy results in a marginal decrease in performance, with our model solving one fewer problem. Thus, it offers substantial savings in training time and computational resources (40\% less training data) with a negligible impact on the final proving capability. Second, when the critic model's score-based guidance is replaced with random selection of nodes for expansion, there is a drastic performance degradation (e.g., from 177 to 164 problems solved). This indicates that the learned critic model is effective in navigating the vast search space and guiding the prover towards promising proof paths. Finally, ablating each of our three heuristic rules for multi-perspective search by replacing their specific selection criteria with random choices also results in noticeable performance drops. In these variations ("w/o Tactic Eff. Heuristic", "w/o Min. Cases Heuristic", "w/o Short. State Heuristic"), the specific heuristic rule is deactivated, and its slot in the multi-perspective selection is filled by a random choice from the $N_{samples}$ candidate next states generated by the LLM for the current expansion. This demonstrates that without these heuristic rules diversifying tactic selection and guiding search, the prover is more susceptible to the inherent biases of the critic model alone. It becomes more likely to fall into local minima, explore unproductive tactic sequences, or even reach unprovable states.

\vspace{-0.25em}
\subsection{Proof Length and Diversity Analysis}
\vspace{-0.25em}
To further investigate the characteristics of the proofs generated by different search strategies, we conduct a quantitative comparison between our Multi-Perspective Search (MPS) and standard Best-First Search (BFS) with tree-based distance prediction as critic. We ensure a fair comparison by using the identical LLM backbone and analyzing only the set of problems successfully proven by \emph{both} MPS and BFS, guaranteeing analysis on the same theorems.

\begin{figure*} 
    \centering
    \begin{subfigure}[b]{0.495\textwidth}
        \centering
        \includegraphics[width=\textwidth]{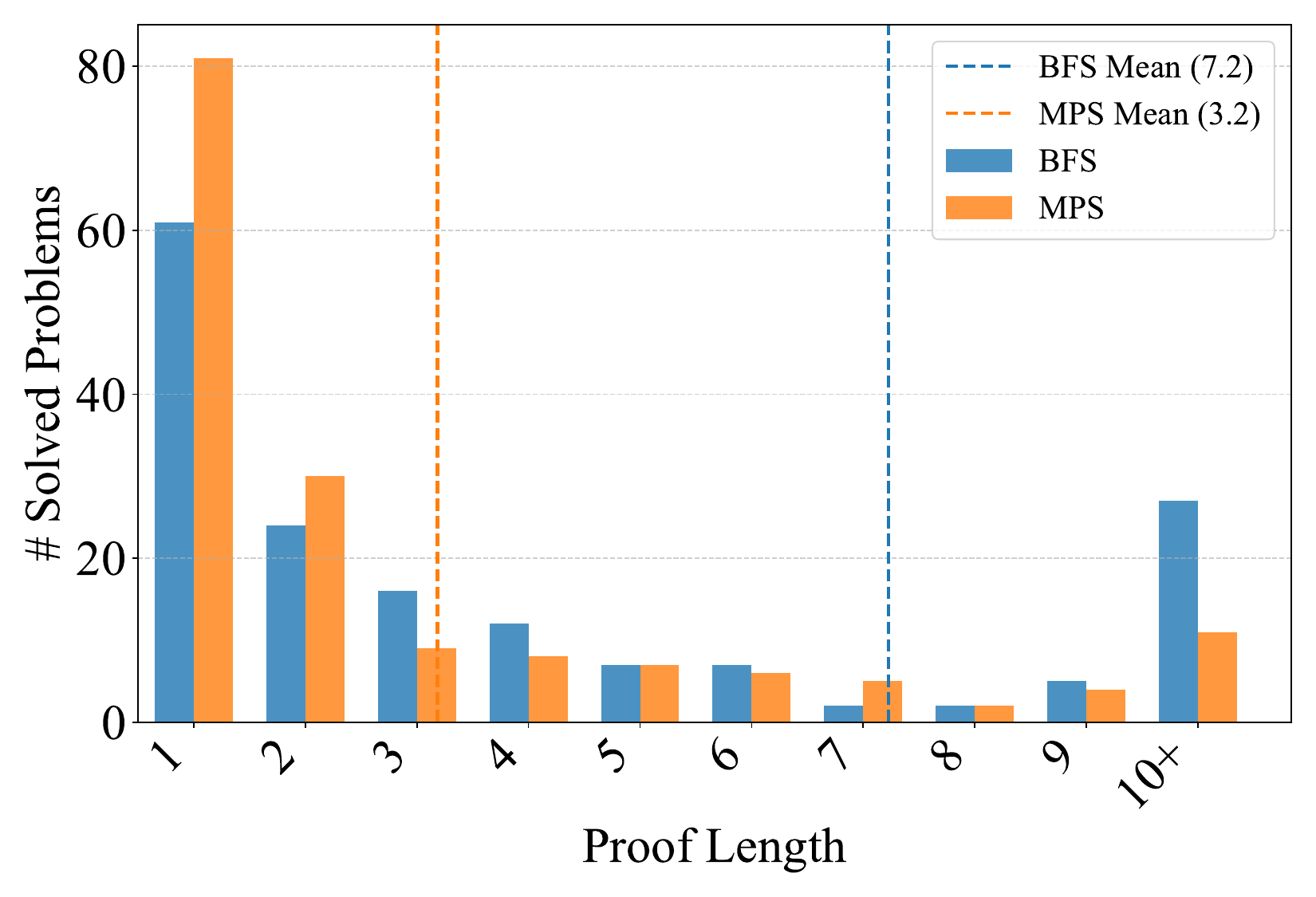}
        \caption{Proof length distribution (steps) for commonly solved problems. Dashed lines indicate mean length.}
        \label{fig:proof_length_side}
    \end{subfigure}
    \hfill 
    \begin{subfigure}[b]{0.495\textwidth} 
        \centering
        \includegraphics[width=\textwidth]{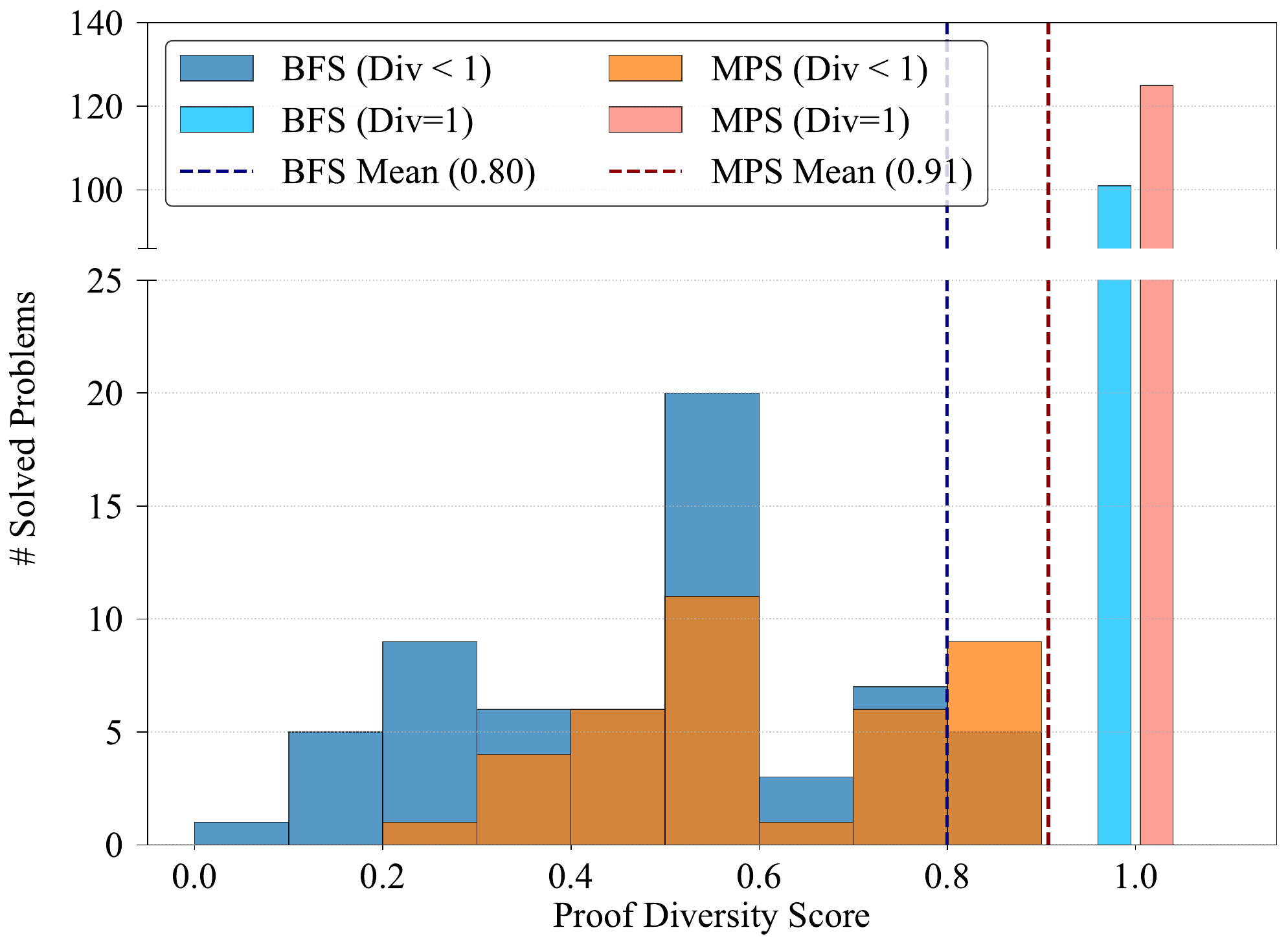}
        \caption{Proof diversity score distribution for commonly solved problems. Dashed lines indicate mean score.}
        \label{fig:proof_diversity_side}
    \end{subfigure}
    \caption{Quantitative analysis of proof characteristics for commonly solved problems by BFS (pass@256) and MPS (pass@64).}
    \vspace{-0.5cm}
    \label{fig:proof_analysis} 
\end{figure*}

Figure~\ref{fig:proof_length_side} shows the distribution of proof lengths, measured as the number of tactic steps (grouped into categories 1-9 and 10+). It is evident that proofs generated by MPS are significantly shorter on average than those found by BFS, as indicated by the mean values (dashed lines). MPS produces a higher frequency of proofs with fewer steps, while BFS exhibits a longer tail of lengthier proofs. This suggests that the diverse guidance signals in MPS help avoid unproductive tactic sequences or local optima, leading to more concise solutions.

Figure~\ref{fig:proof_diversity_side} illustrates the distribution of proof tactic diversity. We define diversity as the number of unique tactics used in a proof divided by its total length (number of steps). A score closer to 1 indicates a wider variety of tactics relative to length. The results clearly show that MPS-generated proofs possess considerably higher average diversity scores compared to BFS proofs (see mean lines). While both methods generate proofs with maximal diversity (score = 1.0, detailed in the annotation), BFS yields a much larger proportion of proofs with very low diversity scores. This highlights MPS's effectiveness in promoting exploration and leveraging a broader range of tactics, whereas BFS, guided solely by the critic model, is more prone to repetitive tactic usage.


\vspace{-0.25em}
\subsection{MPS-Prover vs. Whole-Proof Provers: Proof Length}
\vspace{-0.25em}

\begin{wraptable}{r}{0.5\textwidth}
    \centering
    \vspace{-1.2em} 
    \caption{Proof length statistics (Lean steps) on 170 common miniF2F problems.}
    \label{tab:length_comparison_whole_proof_concise}
    \footnotesize
    \begin{adjustbox}{max width=0.58\textwidth}
    \begin{tabular}{@{}lccc@{}}
    \toprule
    \textbf{Statistic} & \textbf{DeepSeek-V2} & \textbf{Kimina} & \textbf{MPS-Prover} \\
    \midrule
    Min         & 3     & 1     & 1     \\
    Max         & 698   & 186   & 37    \\
    Mean        & 52.16 & 15.91 & 3.44  \\
    Median      & 33.0  & 6.0   & 2.0   \\
    Std Dev     & 66.47 & 24.19 & 4.64  \\
    \bottomrule
    \vspace{-0.6cm}
    \end{tabular}
    \end{adjustbox}
\end{wraptable}

We further analyze proof characteristics by comparing the length of proofs generated by our MPS-Prover against two leading whole-proof provers, Kimina-Prover-Preview and DeepSeek-Prover-V2. This comparison uses 170 commonly solved miniF2F problems and measures proof length in Lean tactic steps. Table~\ref{tab:length_comparison_whole_proof_concise} reveals that MPS-Prover generates substantially shorter proofs (mean length 3.44 steps) compared to Kimina (15.91) and DeepSeek-Prover-V2 (52.16). Some examples of their proofs can be found in our Appendix \ref{app:case_study}.

We attribute this to the operational differences: stepwise provers like MPS-Prover benefit from frequent interactions with the Lean engine. Each tactic execution updates the proof state, allowing the prover to adaptively refine its strategy by treating the new state as a sub-problem. This iterative process, combined with tactic-level search prioritizing impactful steps, facilitates the discovery of more direct solutions. In contrast, whole-proof systems often plan the entire proof initially, with limited dynamic adaptation, potentially leading to longer, albeit correct, proof scripts. We believe this analysis highlights a key advantage of step provers in producing more efficient proofs.

\vspace{-0.25em}
\section{Related Work}
\vspace{-0.25em}
\paragraph{Earlier Methods in Automated Theorem Proving}
Early automated theorem provers relied on symbolic search algorithms and hand-crafted heuristics. Systems like Vampire \citep{riazanov2001vampire} (a first-order logic prover) and SMT solvers like Z3 \citep{de2008proofs} achieved impressive results using resolution, paramodulation, and DPLL-based search without learning. Some researchers apply premise selection in learning. Given a large library of axioms or lemmas, the goal is to predict which ones are relevant to a new theorem. \citet{irving2016deepmath} pioneered deep learning for premise selection with the DeepMath system, using sequence models to rank premises for a target theorem. Similarly, \citet{wang2017premise} used graph embeddings of knowledge bases to select premises, treating theorem proving as a graph traversal problem. Neural networks were also used to guide proof search directly inside automated provers. \citet{loos2017deep} integrated a deep network into the E theorem prover, training the network to predict which inference step to pursue.



\vspace{-0.25em}
\paragraph{LLM-based Whole-Proof Methods.} Recent advances have seen LLMs directly generating complete formal proofs without iterative search, exploiting their powerful sequence modeling capabilities. Early examples include the Draft, Sketch, and Prove (DSP) system \citep{dsp}, which uses informal natural language proofs as guidance, significantly improving prover accuracy; Baldur \citep{first2023baldur} generates proofs for Isabelle theorems and employs a repair mechanism leveraging failure feedback, achieving state-of-the-art results. LEGO-Prover \citep{wang2024lego} enhances whole-proof generation by hierarchically proving and reusing lemmas, effectively managing intermediate results. Similarly, POETRY \citep{wang2024proving} employs recursive proof decomposition, systematically breaking complex theorems into solvable subgoals. Additionally, curriculum learning strategies \citep{polu2022formal} and reinforcement learning \citep{dong2024formal,xin2024deepseek} have been employed to optimize LLM performance. Goedel-Prover \citep{lin2025goedel} and leanabell \citep{zhang2025leanabell} perform continual training with cognitive behavior data and RL outcomes from Lean 4 compiler. Kimina-Prover \citep{wang2025kimina} demonstrates superior results (80.7\% on miniF2F pass@8192) through structured reasoning patterns and RL training. DeepSeek-Prover-V2 \citep{ren2025deepseek} is trained via a recursive subgoal decomposition pipeline using DeepSeek-V3. By integrating CoT-style reasoning with formal proving, it achieves an impressive 88.9\% accuracy on miniF2F.

\vspace{-0.25em}
\paragraph{LLM-based Step-level Tactic Generation Methods.} Stepwise methods integrate LLMs into iterative proof searches, proposing individual proof steps and navigating search trees. GPT-f \citep{polu2020generative} pioneered this approach, proposing tactics that are verified incrementally, laying the groundwork for subsequent systems. HyperTree Proof Search (HTPS) \citep{lample2022hypertree} utilized an AlphaZero-inspired algorithm, exploring multiple proof branches simultaneously, significantly outperforming earlier methods through sophisticated search heuristics. LeanDojo’s ReProver \citep{yang2023leandojo} incorporates premise retrieval, selecting relevant lemmas at each proof step, enhancing efficiency on Lean benchmarks. SubgoalXL \citep{zhao2024subgoalxl} employs expert-guided iterative training, optimizing subgoal generation strategies. ProofAug \citep{liu2025efficient} further develops hybrid integration by alternately invoking neural suggestions, symbolic ATP calls, and recursive prover applications for efficient verification. Recent models like InternLM2.5-StepProver \citep{wu2024internlm2} utilized expert iteration with large-scale datasets.  HunyuanProver \citep{li2024hunyuanprover} further enhanced data synthesis and guided tree search algorithms. BFS-Prover \citep{xin2025bfs} demonstrated the efficacy of simpler Best-First Search methods, incorporating direct preference optimization from compiler feedback, and length normalization.

\vspace{-0.25em}
\section{Discussion}
\vspace{-0.25em}
In this work, we present the Multi-Perspective Search Prover (MPS-Prover), a novel stepwise automated theorem proving system that significantly advances the state of the art. By introducing a principled post-training data curation strategy and a multi-perspective tree search mechanism enhanced with heuristic critics, MPS-Prover effectively addresses common failure modes in existing stepwise provers, such as biased search and exploration of unproductive proof paths. Our extensive experiments demonstrate that MPS-Prover not only achieves superior success rates on challenging benchmarks like miniF2F and ProofNet but also generates proofs that are more concise and diverse. 

For the broader ATP research community, our findings comfirms the strengths of stepwise proving, particularly in generating efficient proofs. Looking ahead, several promising avenues for future work emerge. One key direction is the development of hybrid systems that integrate the global planning capabilities of whole-proof methods with the adaptive, fine-grained search of stepwise provers like MPS-Prover. Another exciting prospect involves combining MPS-Prover with reinforcement learning (RL) techniques to further refine the critic model and search heuristics from self-play or direct feedback from the proof assistant. We believe that these future directions will continue to drive progress towards increasingly powerful and reliable automated theorem proving systems.

\bibliography{colm2024_conference}

\begin{thebibliography}{35}
\providecommand{\natexlab}[1]{#1}
\providecommand{\url}[1]{\texttt{#1}}
\expandafter\ifx\csname urlstyle\endcsname\relax
  \providecommand{\doi}[1]{doi: #1}\else
  \providecommand{\doi}{doi: \begingroup \urlstyle{rm}\Url}\fi

\bibitem[AlphaProof \& Teams(2024)AlphaProof and Teams]{alphaproof2024ai}
DeepMind AlphaProof and AlphaGeometry Teams.
\newblock Ai achieves silver-medal standard solving international mathematical olympiad problems.’25 july 2024, 2024.

\bibitem[Azerbayev et~al.(2023)Azerbayev, Piotrowski, Schoelkopf, Ayers, Radev, and Avigad]{azerbayev2023proofnet}
Zhangir Azerbayev, Bartosz Piotrowski, Hailey Schoelkopf, Edward~W Ayers, Dragomir Radev, and Jeremy Avigad.
\newblock Proofnet: Autoformalizing and formally proving undergraduate-level mathematics.
\newblock \emph{arXiv preprint arXiv:2302.12433}, 2023.

\bibitem[de~Moura \& Bj{\o}rner(2008)de~Moura and Bj{\o}rner]{de2008proofs}
Leonardo~Mendon{\c{c}}a de~Moura and Nikolaj~S Bj{\o}rner.
\newblock Proofs and refutations, and z3.
\newblock In \emph{LPAR Workshops}, volume 418, pp.\  123--132. Doha, Qatar, 2008.

\bibitem[Dong \& Ma(2025)Dong and Ma]{dong2025stp}
Kefan Dong and Tengyu Ma.
\newblock Stp: Self-play llm theorem provers with iterative conjecturing and proving.
\newblock \emph{arXiv e-prints}, pp.\  arXiv--2502, 2025.

\bibitem[Dong et~al.(2024)Dong, Mahankali, and Ma]{dong2024formal}
Kefan Dong, Arvind Mahankali, and Tengyu Ma.
\newblock Formal theorem proving by rewarding llms to decompose proofs hierarchically.
\newblock \emph{arXiv preprint arXiv:2411.01829}, 2024.

\bibitem[First et~al.(2023)First, Rabe, Ringer, and Brun]{first2023baldur}
Emily First, Markus~N Rabe, Talia Ringer, and Yuriy Brun.
\newblock Baldur: Whole-proof generation and repair with large language models.
\newblock In \emph{Proceedings of the 31st ACM Joint European Software Engineering Conference and Symposium on the Foundations of Software Engineering}, pp.\  1229--1241, 2023.

\bibitem[Guo et~al.(2025)Guo, Yang, Zhang, Song, Zhang, Xu, Zhu, Ma, Wang, Bi, et~al.]{guo2025deepseek}
Daya Guo, Dejian Yang, Haowei Zhang, Junxiao Song, Ruoyu Zhang, Runxin Xu, Qihao Zhu, Shirong Ma, Peiyi Wang, Xiao Bi, et~al.
\newblock Deepseek-r1: Incentivizing reasoning capability in llms via reinforcement learning.
\newblock \emph{arXiv preprint arXiv:2501.12948}, 2025.

\bibitem[Irving et~al.(2016)Irving, Szegedy, Alemi, E{\'e}n, Chollet, and Urban]{irving2016deepmath}
Geoffrey Irving, Christian Szegedy, Alexander~A Alemi, Niklas E{\'e}n, Fran{\c{c}}ois Chollet, and Josef Urban.
\newblock Deepmath-deep sequence models for premise selection.
\newblock \emph{Advances in neural information processing systems}, 29, 2016.

\bibitem[Jiang et~al.(2023)Jiang, Welleck, Zhou, Lacroix, Liu, Li, Jamnik, Lample, and Wu]{dsp}
Albert~Qiaochu Jiang, Sean Welleck, Jin~Peng Zhou, Timothee Lacroix, Jiacheng Liu, Wenda Li, Mateja Jamnik, Guillaume Lample, and Yuhuai Wu.
\newblock Draft, sketch, and prove: Guiding formal theorem provers with informal proofs.
\newblock In \emph{The Eleventh International Conference on Learning Representations}, 2023.

\bibitem[Lample et~al.(2022)Lample, Lacroix, Lachaux, Rodriguez, Hayat, Lavril, Ebner, and Martinet]{lample2022hypertree}
Guillaume Lample, Timothee Lacroix, Marie-Anne Lachaux, Aurelien Rodriguez, Amaury Hayat, Thibaut Lavril, Gabriel Ebner, and Xavier Martinet.
\newblock Hypertree proof search for neural theorem proving.
\newblock \emph{Advances in neural information processing systems}, 35:\penalty0 26337--26349, 2022.

\bibitem[Li et~al.(2024{\natexlab{a}})Li, Beeching, Tunstall, Lipkin, Soletskyi, Huang, Rasul, Yu, Jiang, Shen, et~al.]{li2024numinamath}
Jia Li, Edward Beeching, Lewis Tunstall, Ben Lipkin, Roman Soletskyi, Shengyi Huang, Kashif Rasul, Longhui Yu, Albert~Q Jiang, Ziju Shen, et~al.
\newblock Numinamath: The largest public dataset in ai4maths with 860k pairs of competition math problems and solutions.
\newblock \emph{Hugging Face repository}, 13:\penalty0 9, 2024{\natexlab{a}}.

\bibitem[Li et~al.(2024{\natexlab{b}})Li, Du, Song, Li, Wang, Yang, and Mi]{li2024hunyuanprover}
Yang Li, Dong Du, Linfeng Song, Chen Li, Weikang Wang, Tao Yang, and Haitao Mi.
\newblock Hunyuanprover: A scalable data synthesis framework and guided tree search for automated theorem proving.
\newblock \emph{arXiv preprint arXiv:2412.20735}, 2024{\natexlab{b}}.

\bibitem[Lin et~al.(2025)Lin, Tang, Lyu, Wu, Lin, Yang, Li, Xia, Chen, Arora, et~al.]{lin2025goedel}
Yong Lin, Shange Tang, Bohan Lyu, Jiayun Wu, Hongzhou Lin, Kaiyu Yang, Jia Li, Mengzhou Xia, Danqi Chen, Sanjeev Arora, et~al.
\newblock Goedel-prover: A frontier model for open-source automated theorem proving.
\newblock \emph{arXiv preprint arXiv:2502.07640}, 2025.

\bibitem[Liu et~al.(2025)Liu, Sun, Li, and Yao]{liu2025efficient}
Haoxiong Liu, Jiacheng Sun, Zhenguo Li, and Andrew~C Yao.
\newblock Efficient neural theorem proving via fine-grained proof structure analysis.
\newblock \emph{arXiv preprint arXiv:2501.18310}, 2025.

\bibitem[Loos et~al.(2017)Loos, Irving, Szegedy, and Kaliszyk]{loos2017deep}
Sarah Loos, Geoffrey Irving, Christian Szegedy, and Cezary Kaliszyk.
\newblock Deep network guided proof search.
\newblock \emph{arXiv preprint arXiv:1701.06972}, 2017.

\bibitem[Mahdavi et~al.(2025)Mahdavi, Li, Liu, Thrampoulidis, Sigal, and Liao]{mahdavi2025leveraging}
Sadegh Mahdavi, Muchen Li, Kaiwen Liu, Christos Thrampoulidis, Leonid Sigal, and Renjie Liao.
\newblock Leveraging online olympiad-level math problems for llms training and contamination-resistant evaluation.
\newblock \emph{arXiv preprint arXiv:2501.14275}, 2025.

\bibitem[Moura \& Ullrich(2021)Moura and Ullrich]{moura2021lean}
Leonardo~de Moura and Sebastian Ullrich.
\newblock The lean 4 theorem prover and programming language.
\newblock In \emph{Automated Deduction--CADE 28: 28th International Conference on Automated Deduction, Virtual Event, July 12--15, 2021, Proceedings 28}, pp.\  625--635. Springer, 2021.

\bibitem[Paulson(1994)]{paulson1994isabelle}
Lawrence~C Paulson.
\newblock \emph{Isabelle: A generic theorem prover}.
\newblock Springer, 1994.

\bibitem[Polu \& Sutskever(2020)Polu and Sutskever]{polu2020generative}
Stanislas Polu and Ilya Sutskever.
\newblock Generative language modeling for automated theorem proving.
\newblock \emph{arXiv preprint arXiv:2009.03393}, 2020.

\bibitem[Polu et~al.(2022)Polu, Han, Zheng, Baksys, Babuschkin, and Sutskever]{polu2022formal}
Stanislas Polu, Jesse~Michael Han, Kunhao Zheng, Mantas Baksys, Igor Babuschkin, and Ilya Sutskever.
\newblock Formal mathematics statement curriculum learning.
\newblock \emph{arXiv preprint arXiv:2202.01344}, 2022.

\bibitem[Ren et~al.(2025)Ren, Shao, Song, Xin, Wang, Zhao, Zhang, Fu, Zhu, Yang, et~al.]{ren2025deepseek}
ZZ~Ren, Zhihong Shao, Junxiao Song, Huajian Xin, Haocheng Wang, Wanjia Zhao, Liyue Zhang, Zhe Fu, Qihao Zhu, Dejian Yang, et~al.
\newblock Deepseek-prover-v2: Advancing formal mathematical reasoning via reinforcement learning for subgoal decomposition.
\newblock \emph{arXiv preprint arXiv:2504.21801}, 2025.

\bibitem[Riazanov \& Voronkov(2001)Riazanov and Voronkov]{riazanov2001vampire}
Alexandre Riazanov and Andrei Voronkov.
\newblock Vampire 1.1.
\newblock In \emph{Automated Reasoning: First International Joint Conference, IJCAR 2001 Siena, Italy, June 18--22, 2001 Proceedings 1}, pp.\  376--380. Springer, 2001.

\bibitem[Wang et~al.(2024{\natexlab{a}})Wang, Xin, Liu, Li, Huang, Lu, Yang, Tang, Yin, Li, et~al.]{wang2024proving}
Haiming Wang, Huajian Xin, Zhengying Liu, Wenda Li, Yinya Huang, Jianqiao Lu, Zhicheng Yang, Jing Tang, Jian Yin, Zhenguo Li, et~al.
\newblock Proving theorems recursively.
\newblock \emph{arXiv preprint arXiv:2405.14414}, 2024{\natexlab{a}}.

\bibitem[Wang et~al.(2024{\natexlab{b}})Wang, Xin, Zheng, Liu, Cao, Huang, Xiong, Shi, Xie, Yin, et~al.]{wang2024lego}
Haiming Wang, Huajian Xin, Chuanyang Zheng, Zhengying Liu, Qingxing Cao, Yinya Huang, Jing Xiong, Han Shi, Enze Xie, Jian Yin, et~al.
\newblock Lego-prover: Neural theorem proving with growing libraries.
\newblock In \emph{12th International Conference on Learning Representations (ICLR 2024)}. International Conference on Learning Representations, ICLR, 2024{\natexlab{b}}.

\bibitem[Wang et~al.(2025)Wang, Unsal, Lin, Baksys, Liu, Santos, Sung, Vinyes, Ying, Zhu, et~al.]{wang2025kimina}
Haiming Wang, Mert Unsal, Xiaohan Lin, Mantas Baksys, Junqi Liu, Marco~Dos Santos, Flood Sung, Marina Vinyes, Zhenzhe Ying, Zekai Zhu, et~al.
\newblock Kimina-prover preview: Towards large formal reasoning models with reinforcement learning.
\newblock \emph{arXiv preprint arXiv:2504.11354}, 2025.

\bibitem[Wang et~al.(2017)Wang, Tang, Wang, and Deng]{wang2017premise}
Mingzhe Wang, Yihe Tang, Jian Wang, and Jia Deng.
\newblock Premise selection for theorem proving by deep graph embedding.
\newblock \emph{Advances in neural information processing systems}, 30, 2017.

\bibitem[Wang et~al.(2024{\natexlab{c}})Wang, Zhang, Jia, Pan, Diao, Pi, and Zhang]{wang2024theoremllama}
Ruida Wang, Jipeng Zhang, Yizhen Jia, Rui Pan, Shizhe Diao, Renjie Pi, and Tong Zhang.
\newblock Theoremllama: Transforming general-purpose llms into lean4 experts.
\newblock In \emph{Proceedings of the 2024 Conference on Empirical Methods in Natural Language Processing}, pp.\  11953--11974, 2024{\natexlab{c}}.

\bibitem[Wu et~al.(2024)Wu, Huang, Zhou, Ying, Wang, Lin, and Chen]{wu2024internlm2}
Zijian Wu, Suozhi Huang, Zhejian Zhou, Huaiyuan Ying, Jiayu Wang, Dahua Lin, and Kai Chen.
\newblock Internlm2. 5-stepprover: Advancing automated theorem proving via expert iteration on large-scale lean problems.
\newblock \emph{arXiv preprint arXiv:2410.15700}, 2024.

\bibitem[Xin et~al.(2024)Xin, Ren, Song, Shao, Zhao, Wang, Liu, Zhang, Lu, Du, et~al.]{xin2024deepseek}
Huajian Xin, ZZ~Ren, Junxiao Song, Zhihong Shao, Wanjia Zhao, Haocheng Wang, Bo~Liu, Liyue Zhang, Xuan Lu, Qiushi Du, et~al.
\newblock Deepseek-prover-v1. 5: Harnessing proof assistant feedback for reinforcement learning and monte-carlo tree search.
\newblock \emph{arXiv preprint arXiv:2408.08152}, 2024.

\bibitem[Xin et~al.(2025)Xin, Xi, Yang, Chen, Wu, Xiao, Sun, Zheng, and Shen]{xin2025bfs}
Ran Xin, Chenguang Xi, Jie Yang, Feng Chen, Hang Wu, Xia Xiao, Yifan Sun, Shen Zheng, and Kai Shen.
\newblock Bfs-prover: Scalable best-first tree search for llm-based automatic theorem proving.
\newblock \emph{arXiv preprint arXiv:2502.03438}, 2025.

\bibitem[Yang et~al.(2023)Yang, Swope, Gu, Chalamala, Song, Yu, Godil, Prenger, and Anandkumar]{yang2023leandojo}
Kaiyu Yang, Aidan Swope, Alex Gu, Rahul Chalamala, Peiyang Song, Shixing Yu, Saad Godil, Ryan~J Prenger, and Animashree Anandkumar.
\newblock Leandojo: Theorem proving with retrieval-augmented language models.
\newblock \emph{Advances in Neural Information Processing Systems}, 36:\penalty0 21573--21612, 2023.

\bibitem[Ying et~al.(2024)Ying, Wu, Geng, Wang, Lin, and Chen]{yinglean}
Huaiyuan Ying, Zijian Wu, Yihan Geng, JIayu Wang, Dahua Lin, and Kai Chen.
\newblock Lean workbook: A large-scale lean problem set formalized from natural language math problems.
\newblock In \emph{The Thirty-eight Conference on Neural Information Processing Systems Datasets and Benchmarks Track}, 2024.

\bibitem[Zhang et~al.(2025)Zhang, Wang, Ji, Liu, Yue, Zhang, Zhang, Zhou, and Gai]{zhang2025leanabell}
Jingyuan Zhang, Qi~Wang, Xingguang Ji, Yahui Liu, Yang Yue, Fuzheng Zhang, Di~Zhang, Guorui Zhou, and Kun Gai.
\newblock Leanabell-prover: Posttraining scaling in formal reasoning.
\newblock \emph{arXiv preprint arXiv:2504.06122}, 2025.

\bibitem[Zhao et~al.(2024)Zhao, Zheng, Bo, Hu, Thakker, and Kong]{zhao2024subgoalxl}
Xueliang Zhao, Lin Zheng, Haige Bo, Changran Hu, Urmish Thakker, and Lingpeng Kong.
\newblock Subgoalxl: Subgoal-based expert learning for theorem proving.
\newblock \emph{arXiv preprint arXiv:2408.11172}, 2024.

\bibitem[Zheng et~al.(2022)Zheng, Han, and Polu]{zheng2022miniF2F}
Kunhao Zheng, Jesse~Michael Han, and Stanislas Polu.
\newblock minif2f: a cross-system benchmark for formal olympiad-level mathematics.
\newblock In \emph{ICLR}, 2022.

\end{thebibliography}
\bibliographystyle{colm2024_conference}

\appendix
\section{Limitation}

While our MPS-Prover demonstrates significant advancements in stepwise automated theorem proving, it is important to acknowledge certain inherent limitations of the stepwise paradigm itself, particularly when compared to whole-proof generation approaches.

A primary limitation of current stepwise provers, including MPS-Prover, lies in their handling of tactics that introduce complex, nested proof obligations, such as new lemmas that require their own sub-proofs. Whole-proof systems, like DeepSeek-Prover-V2 \citep{guo2025deepseek}, can generate entire proof scripts that include $have$ statements to introduce and subsequently prove auxiliary lemmas within the main proof structure. The verifier then processes the complete script. However, for a purely stepwise prover interacting with the Lean 4 engine, if a tactic attempts to introduce an unproven lemma or a complex structure requiring an immediate, unfulfilled sub-proof (e.g., via $have$, or certain intricate uses of $induction$ or $calc$ blocks that don't immediately resolve to simpler goals), the Lean engine will typically raise an error and halt that proof path. The prover cannot easily "pause" the main proof, prove the lemma in isolation, and then resume, all within a single interactive step.

This means that MPS-Prover, like other current step-provers, is less adept at autonomously discovering and utilizing complex intermediate lemmas that are not already present in the context or standard libraries. While our multi-perspective search can find efficient paths using existing tactics, it does not inherently support the generation and in-line proving of new, non-trivial lemmas in the same way a whole-proof generator might plan for. This restricts the prover's ability to break down very complex problems into more manageable, lemma-dependent sub-problems in a self-contained manner during the stepwise search.

Addressing this limitation is a key direction for future work. As mentioned in our conclusion, exploring hybrid approaches that combine the stepwise search capabilities of MPS-Prover with the global planning and lemma-handling strengths of whole-proof generation methods could offer a promising path towards overcoming this challenge and further expanding the scope of theorems that can be automatically proven.

\section{Tactic Effectiveness Scoring}
\label{app:score}
The Tactic Effectiveness Scoring heuristic in MPS-Prover assigns a numerical score to potential next steps based on the tactic used. These scores are designed to prioritize tactics that are generally more impactful or transformative in the proof process, while giving lower scores to auxiliary or very general-purpose tactics that might be applied speculatively. The scores are based on common patterns observed in mathematical proofs and expert experience with the Lean theorem prover. The goal is to guide the search towards more direct and structured proofs by favoring steps that represent significant logical advancements.

Below is the scoring table used. Tactics are grouped by score, with higher scores indicating a stronger preference. Note that regular expressions are used for some tactic patterns (e.g., $simp??$ matches $simp$ and $simp?$).

\begin{table}[h!]
\centering
\caption{Tactic Effectiveness Scores}
\label{tab:tactic_scores}
\begin{tabular}{cl}
\toprule
\textbf{Score} & \textbf{Tactics / Patterns} \\
\midrule
6 & \texttt{exact}, \texttt{refine}, \texttt{rintro}, \texttt{rcases}, \texttt{induction}, \texttt{revert}, \\
  & \texttt{by\_contra}, \texttt{contrapose} \\
\midrule
5 & \texttt{rw}, \texttt{rw .* at}, \texttt{convert}, \texttt{apply}, \texttt{subst}, \texttt{linarith}, \\
  & \texttt{congr}, \texttt{ring\_nf} \\
\midrule
4 & \texttt{ring}, \texttt{field\_simp}, \texttt{group}, \texttt{aesop} \\
\midrule
3 & \texttt{simp??}, \texttt{simp\_all}, \texttt{simp only} \\
\midrule
2 & \texttt{norm\_cast}, \texttt{push\_cast}, \texttt{clear} \\
\midrule
1 & \texttt{norm\_num}, \texttt{swap}, \texttt{all\_goals} \\
\midrule
0 & \texttt{have x = y} (without a subsequent $by$ block for proof) \\
\bottomrule
\end{tabular}
\end{table}

Below is the rationale for Scoring Tiers:

\paragraph{Score 6 (Highly Transformative/Goal-Closing):} This tier includes tactics that often conclude a proof branch directly (e.g., $exact$, $refine$), introduce crucial case distinctions or structural changes (e.g., $induction$, $rcases$), or fundamentally alter the goal's form (e.g., $by_contra$, $contrapose$, $revert$). These are typically strong indicators of significant progress.

\paragraph{Score 5 (Strong Rewriting/Application):} Tactics like $rw$ (rewrite), $apply$ (apply a hypothesis/lemma), and $convert$ (change goal to a definitionally equal one) are powerful for making targeted changes. Algebraic simplification tools like $linarith$ and $ring\_nf$ also fall here as they can often solve subgoals involving arithmetic or ring structures.

\paragraph{Score 4 (Domain-Specific Solvers/Automated Tactics):} This includes more specialized solvers like $ring$ (for ring equalities) and $field\_simp$ (for field simplifications), $group$ (for group theory), as well as general automated tactics like $aesop$. While powerful, $aesop$ is placed slightly lower than the top tier as it can sometimes be a "black box" and its success is highly conditional.

\paragraph{Score 3 (General Simplification):} The $simp$ family of tactics ($simp$, $simp\_all$, $simp only$) are general-purpose simplification tools. They are very useful but are scored moderately because they can sometimes be applied excessively or ineffectively, leading to many unproductive steps. For tactics that are excluded in the table, we assign a default score of 3.

\paragraph{Score 2 (Normalization/Cleanup):} Tactics like $norm\_cast$ (normalize casts), $push\_cast$ (push casts inwards/outwards), and $clear$ (remove unused hypotheses) are important for maintaining a clean and manageable proof state but don't usually represent major logical steps forward.

\paragraph{Score 1 (Auxiliary/Low Impact):} This includes very basic numerical normalization ($norm_num$), reordering goals ($swap$), or meta-level tactic combinators ($all\_goals$) which are generally supportive rather than primary drivers of proof progress.

\paragraph{Score 0 (Potentially Redundant $have$):} A $have x = y$ statement without an accompanying $by$ block to prove it (implying it might be proved by a trivial step, or is simply a renaming) is given the lowest score. The intent here is to penalize the simple act of stating a trivial equality without further justification as a standalone step.

This heuristic scoring aims to complement the learned critic model by providing a stable, experience-based bias towards tactics that are historically effective in structuring and advancing mathematical proofs.

\section{Baselines}
\label{app:baseline}
We compare MPS-Prover against a comprehensive set of state-of-the-art automated theorem provers. For whole-proof generation methods, we include Kimina-Prover-Preview \citep{wang2025kimina}, which employs interleaved natural language reasoning and Lean code blocks along with reinforcement learning (RL). Another strong contender is DeepSeek-prover V2 \citep{ren2025deepseek}, notable for its use of subgoal decomposition to break down complex problems and subsequent proof generation, also enhanced with RL. Goedel prover \citep{lin2025goedel} represents methods focused on extensive data collection, having curated a large formalized mathematics dataset for expert iteration training. Leanabell-Prover \citep{zhang2025leanabell} similarly combines expert iteration with RL techniques. Additionally, STP \citep{dong2025stp} utilizes a dual-role architecture with a conjecturer and a prover, where each component provides training signals for the other. For stepwise proof generation methods, we select InternLM-StepProver \citep{wu2024internlm2}, one of the pioneers in applying LLMs to step-level ATP. Hunyuan prover \citep{li2024hunyuanprover} advanced this line by designing improved critic models and integrating Monte Carlo Tree Search (MCTS). The most recent and leading baseline in this category is BFS-prover \citep{xin2025bfs}, which combines Supervised Fine-tuning (SFT) with Direct Preference Optimization (DPO) and incorporates length normalization during its Best-First Search, representing the previous state-of-the-art for step-provers.

\section{Case Studies}
\label{app:case_study}
To provide a more nuanced understanding of the differences in proof strategies and generated solutions, we conduct case studies on specific theorems. We compare proofs generated by our MPS-Prover with those from Kimina-Prover and DeepSeek-Prover V2 for two commonly solved problems, and additionally showcase a problem uniquely solved by MPS-Prover.

\subsection{Analysis of a Commonly Solved Problem: \texttt{algebra\_absapbon1pabsapbleqsumabsaon1pabsa}}

The theorem \texttt{algebra\_absapbon1pabsapbleqsumabsaon1pabsa} states that for any real numbers $a$ and $b$, $\frac{|a+b|}{1+|a+b|} \leq \frac{|a|}{1+|a|} + \frac{|b|}{1+|b|}$. All three provers successfully found a proof for this theorem, but their approaches and the resulting proof scripts differ significantly in length and style.

Our MPS-Prover generates a remarkably concise proof of only 8 lines. Key steps include leveraging $rw$ for rewriting goals based on non-negativity, using $by\_cases$ for case analysis (e.g., $a=0$), and then efficiently using $field\_simp$ with relevant hypotheses like $abs\_nonneg$ and $mul\_nonneg$. The proof concludes with a call to $refine'$ combined with $div\_nonneg$ and powerful finishers like $nlinarith$ and $positivity$. Each step appears to make substantial progress, often simplifying the goal significantly or discharging parts of it by effectively utilizing built-in Mathlib lemmas and tactics. This conciseness stems from MPS-Prover's ability to explore and select tactics that yield significant progress at each step, guided by the multi-perspective search.

In contrast, the solution from Kimina-Prover for the same problem is considerably longer, spanning approximately 40 lines. It primarily relies on a sequence of $have$ statements to introduce intermediate lemmas (e.g., $h1: abs (a + b) \leq abs a + abs b$, $h2: \dots$, $h3: \dots$). Each of these lemmas is then proven using a combination of more granular tactics like $apply$, $linarith$, $nlinarith$, and $field\_simp$. While logically sound, this approach of explicitly stating and proving multiple intermediate steps results in a more verbose proof. This suggests a strategy that might break down the problem based on an initial plan but does not necessarily find the most direct path available through more powerful, combined tactics.

The proof generated by DeepSeek-Prover V2 is the longest, around 60 lines. Similar to Kimina-Prover, it also extensively uses $have$ to introduce and prove a series of lemmas. However, the proof of those lemmas, and its subsequent applications, are broken down into many fine-grained steps involving $positivity$, $nlinarith$, $linarith$, and $rw [div\_le\_div\_iff \dots]$. While demonstrating a sophisticated understanding by identifying and proving this general property, the overall proof becomes quite lengthy due to the detailed, step-by-step derivation of each component. This is characteristic of whole-proof systems that might formulate a high-level plan (e.g., "prove and use the monotonicity of $f(x)=x/(1+x)$") and then meticulously fill in the details, which may not always be the most compact way to express the proof in Lean.

These comparisons highlight that MPS-Prover, through its stepwise search and selection of high-progress tactics, tends to produce more concise proofs by effectively leveraging powerful built-in Mathlib functionalities. Whole-proof systems, while capable of complex reasoning and planning, may generate longer proofs due to their tendency to break down problems into many simpler, explicitly stated intermediate steps, potentially constrained by an initial, possibly more verbose, plan.

\begin{minted}[fontsize=\small, breaklines, linenos, bgcolor=gray!10]{lean}
-- Solution to algebra_absapbon1pabsapbleqsumabsaon1pabsa by our MPS-Prover

import Mathlib
import Aesop
set_option maxHeartbeats 0
open BigOperators Real Nat Topology Rat

theorem algebra_absapbon1pabsapbleqsumabsaon1pabsa (a b : ℝ) :
    abs (a + b) / (1 + abs (a + b)) ≤ abs a / (1 + abs a) + abs b / (1 + abs b) := by 

  rw [← sub_nonneg, ← sub_nonneg, ← sub_nonneg]
  by_cases h₁ : a = 0
  all_goals field_simp [abs_of_nonneg, abs_of_pos, *]
  have h₂ : 0 ≤ |b| := abs_nonneg b
<;> have h₃ : 0 ≤ |a| := abs_nonneg a
<;> field_simp [h₁, h₂, h₃]
  have h₄ : 0 ≤ |a + b| := abs_nonneg (a + b)
<;>
have h₅ : 0 ≤ |a| * |b| := mul_nonneg h₃ h₂
<;>
field_simp [h₁, h₂, h₃, h₄, h₅]
  refine' div_nonneg (by nlinarith [abs_mul a b, abs_add a b]) (by positivity)
\end{minted}

\begin{minted}[fontsize=\small, breaklines, linenos, bgcolor=gray!10]{lean}
-- Solution to algebra_absapbon1pabsapbleqsumabsaon1pabsa by Kimina-Prover

import Mathlib
import Aesop

set_option maxHeartbeats 0

open BigOperators Real Nat Topology Rat

theorem algebra_absapbon1pabsapbleqsumabsaon1pabsa (a b : ℝ) :
    abs (a + b) / (1 + abs (a + b)) ≤ abs a / (1 + abs a) + abs b / (1 + abs b) := by
  have h1 : abs (a + b) ≤ abs a + abs b := by
    apply abs_add
  have h2 : abs (a + b) / (1 + abs (a + b)) ≤ (abs a + abs b) / (1 + abs a + abs b) := by
    have h2a : abs (a + b) ≤ abs a + abs b := by
      apply abs_add
    have h2b : 0 ≤ abs (a + b) := abs_nonneg (a + b)
    have h2c : 0 ≤ abs a + abs b := by
      apply add_nonneg
      · apply abs_nonneg a
      · apply abs_nonneg b
    have h2d : 0 ≤ (1 + abs (a + b)) := by
      linarith [abs_nonneg (a + b)]
    have h2e : 0 ≤ (1 + abs a + abs b) := by
      linarith [abs_nonneg a, abs_nonneg b]
    have h2f : abs (a + b) / (1 + abs (a + b)) ≤ (abs a + abs b) / (1 + abs a + abs b) := by
      apply (div_le_div_iff (by linarith) (by linarith)).mpr
      nlinarith [abs_nonneg (a + b), abs_nonneg a, abs_nonneg b, h2a]
    exact h2f
  have h3 : (abs a + abs b) / (1 + abs a + abs b) ≤ abs a / (1 + abs a) + abs b / (1 + abs b) := by
    have h3a : 0 ≤ (1 + abs a + abs b) := by
      linarith [abs_nonneg a, abs_nonneg b]
    have h3b : 0 ≤ (1 + abs a) := by
      linarith [abs_nonneg a]
    have h3c : 0 ≤ (1 + abs b) := by
      linarith [abs_nonneg b]
    have h4 : (abs a + abs b) / (1 + abs a + abs b) ≤ abs a / (1 + abs a) + abs b / (1 + abs b) := by
      have h4a : abs a ≥ 0 := abs_nonneg a
      have h4b : abs b ≥ 0 := abs_nonneg b
      have h4c : 1 + abs a + abs b > 0 := by linarith [abs_nonneg a, abs_nonneg b]
      have h4d : 1 + abs a > 0 := by linarith [abs_nonneg a]
      have h4e : 1 + abs b > 0 := by linarith [abs_nonneg b]
      have h5 : abs a / (1 + abs a) + abs b / (1 + abs b) - (abs a + abs b) / (1 + abs a + abs b) ≥ 0 := by
        have h5a : abs a / (1 + abs a) + abs b / (1 + abs b) - (abs a + abs b) / (1 + abs a + abs b) = 
          (abs a * (1 + abs b) * (1 + abs a + abs b) + 
          abs b * (1 + abs a) * (1 + abs a + abs b) - 
          (abs a + abs b) * (1 + abs a) * (1 + abs b)) / 
          ((1 + abs a) * (1 + abs b) * (1 + abs a + abs b)) := by
          field_simp
          <;> ring
        rw [h5a]
        have h5b : 0 ≤ (abs a * (1 + abs b) * (1 + abs a + abs b) + 
          abs b * (1 + abs a) * (1 + abs a + abs b) - 
          (abs a + abs b) * (1 + abs a) * (1 + abs b)) := by 
          nlinarith [sq_nonneg (abs a), sq_nonneg (abs b), mul_nonneg (abs_nonneg a) (abs_nonneg b)]
        have h5c : 0 < (1 + abs a) * (1 + abs b) * (1 + abs a + abs b) := by
          apply mul_pos
          apply mul_pos
          · linarith [abs_nonneg a]
          · linarith [abs_nonneg b]
          · linarith [abs_nonneg a, abs_nonneg b]
        apply div_nonneg
        · linarith
        · nlinarith [abs_nonneg a, abs_nonneg b]
      linarith [h5]
    linarith [h4]
  linarith [h2, h3]
\end{minted}

\begin{minted}[fontsize=\small, breaklines, linenos, bgcolor=gray!10]{lean}
-- Solution to algebra_absapbon1pabsapbleqsumabsaon1pabsa by Deepseek-Prover V2

import Mathlib
import Aesop

set_option maxHeartbeats 0

open BigOperators Real Nat Topology Rat

/-- Show that for any real numbers $a$ and $b$, $\frac{|a+b|}{1+|a+b|}\leq \frac{|a|}{1+|a|}+\frac{|b|}{1+|b|}$.-/
theorem algebra_absapbon1pabsapbleqsumabsaon1pabsa (a b : ℝ) :
    abs (a + b) / (1 + abs (a + b)) ≤ abs a / (1 + abs a) + abs b / (1 + abs b) := by
  have h₁ : abs (a + b) ≤ abs a + abs b := by
    calc
      abs (a + b) ≤ abs a + abs b := by
        -- Use the triangle inequality for absolute values.
        exact abs_add a b
      _ = abs a + abs b := by rfl
  
  have h₂ : ∀ (x y : ℝ), 0 ≤ x → x ≤ y → x / (1 + x) ≤ y / (1 + y) := by
    intro x y hx hxy
    have h₃ : 0 ≤ x := hx
    have h₄ : x ≤ y := hxy
    have h₅ : 0 ≤ y := by linarith
    have h₆ : 0 ≤ 1 + x := by linarith
    have h₇ : 0 ≤ 1 + y := by linarith
    have h₈ : 0 < 1 + x := by linarith
    have h₉ : 0 < 1 + y := by linarith
    -- Use the division inequality to compare the fractions.
    have h₁₀ : x / (1 + x) ≤ y / (1 + y) := by
      rw [div_le_div_iff (by positivity) (by positivity)]
      -- Simplify the inequality to a form that can be handled by linear arithmetic.
      nlinarith [mul_nonneg h₃ (sub_nonneg.mpr h₄), mul_nonneg h₅ h₃,
        mul_nonneg h₅ (sub_nonneg.mpr h₄)]
    exact h₁₀
  
  have h₃ : abs (a + b) / (1 + abs (a + b)) ≤ (abs a + abs b) / (1 + (abs a + abs b)) := by
    have h₄ : 0 ≤ abs (a + b) := abs_nonneg (a + b)
    have h₅ : 0 ≤ abs a + abs b := by positivity
    have h₆ : abs (a + b) ≤ abs a + abs b := h₁
    have h₇ : 0 ≤ 1 + abs (a + b) := by positivity
    have h₈ : 0 ≤ 1 + (abs a + abs b) := by positivity
    have h₉ : 0 < 1 + abs (a + b) := by positivity
    have h₁₀ : 0 < 1 + (abs a + abs b) := by positivity
    -- Use the fact that the function x / (1 + x) is increasing for x ≥ 0.
    have h₁₁ : abs (a + b) / (1 + abs (a + b)) ≤ (abs a + abs b) / (1 + (abs a + abs b)) := by
      have h₁₂ : abs (a + b) / (1 + abs (a + b)) ≤ (abs a + abs b) / (1 + (abs a + abs b)) := by
        -- Apply the increasing property of the function.
        have h₁₃ : 0 ≤ abs (a + b) := by positivity
        have h₁₄ : abs (a + b) ≤ abs a + abs b := h₁
        have h₁₅ : 0 ≤ abs a + abs b := by positivity
        have h₁₆ : 0 < 1 + abs (a + b) := by positivity
        have h₁₇ : 0 < 1 + (abs a + abs b) := by positivity
        -- Use the previously proven lemma.
        have h₁₈ : abs (a + b) / (1 + abs (a + b)) ≤ (abs a + abs b) / (1 + (abs a + abs b)) := by
          apply h₂
          <;> try norm_num
          <;> linarith
        exact h₁₈
      exact h₁₂
    exact h₁₁
  
  have h₄ : (abs a + abs b) / (1 + (abs a + abs b)) ≤ abs a / (1 + abs a) + abs b / (1 + abs b) := by
    have h₅ : 0 ≤ abs a := abs_nonneg a
    have h₆ : 0 ≤ abs b := abs_nonneg b
    have h₇ : 0 ≤ abs a + abs b := by positivity
    have h₈ : 0 < 1 + abs a := by positivity
    have h₉ : 0 < 1 + abs b := by positivity
    have h₁₀ : 0 < 1 + (abs a + abs b) := by positivity
    have h₁₁ : (abs a + abs b) / (1 + (abs a + abs b)) = abs a / (1 + (abs a + abs b)) + abs b / (1 + (abs a + abs b)) := by
      field_simp [h₈, h₉, h₁₀]
      <;> ring
      <;> field_simp [h₈, h₉, h₁₀]
      <;> ring
    rw [h₁₁]
    have h₁₂ : abs a / (1 + (abs a + abs b)) ≤ abs a / (1 + abs a) := by
      have h₁₃ : 0 ≤ abs a := abs_nonneg a
      have h₁₄ : 0 < 1 + abs a := by positivity
      have h₁₅ : 0 < 1 + (abs a + abs b) := by positivity
      have h₁₆ : 0 < 1 + abs a + abs b := by positivity
      -- Use the fact that 1 + abs a ≤ 1 + (abs a + abs b) to show the inequality
      have h₁₇ : 1 + abs a ≤ 1 + (abs a + abs b) := by
        nlinarith [h₆]
      -- Use the division inequality to prove the result
      have h₁₈ : abs a / (1 + (abs a + abs b)) ≤ abs a / (1 + abs a) := by
        apply div_le_div_of_le_left (by positivity) (by positivity)
        <;> nlinarith
      exact h₁₈
    have h₁₃ : abs b / (1 + (abs a + abs b)) ≤ abs b / (1 + abs b) := by
      have h₁₄ : 0 ≤ abs b := abs_nonneg b
      have h₁₅ : 0 < 1 + abs b := by positivity
      have h₁₆ : 0 < 1 + (abs a + abs b) := by positivity
      have h₁₇ : 0 < 1 + abs a + abs b := by positivity
      -- Use the fact that 1 + abs b ≤ 1 + (abs a + abs b) to show the inequality
      have h₁₈ : 1 + abs b ≤ 1 + (abs a + abs b) := by
        nlinarith [h₅]
      -- Use the division inequality to prove the result
      have h₁₉ : abs b / (1 + (abs a + abs b)) ≤ abs b / (1 + abs b) := by
        apply div_le_div_of_le_left (by positivity) (by positivity)
        <;> nlinarith
      exact h₁₉
    have h₂₀ : abs a / (1 + (abs a + abs b)) + abs b / (1 + (abs a + abs b)) ≤ abs a / (1 + abs a) + abs b / (1 + abs b) := by
      linarith
    linarith
  
  have h₅ : abs (a + b) / (1 + abs (a + b)) ≤ abs a / (1 + abs a) + abs b / (1 + abs b) := by
    calc
      abs (a + b) / (1 + abs (a + b)) ≤ (abs a + abs b) / (1 + (abs a + abs b)) := h₃
      _ ≤ abs a / (1 + abs a) + abs b / (1 + abs b) := h₄
  
  exact h₅
\end{minted}

\subsection{Analysis of a Commonly Solved Problem: \texttt{imo\_1962\_p2}}
The theorem \texttt{imo\_1962\_p2} is another problem successfully solved by all three provers, allowing for a comparison of proof styles for a different type of problem involving inequalities and square roots.
Our MPS-Prover again produces a very short proof. It effectively uses $refine'$ to structure the proof for the conjunction, then leverages a sequence of powerful rewriting and simplification tactics like $rw$, $field\_simp$, $norm\_num$, and $nlinarith$, often chained or applied with specific hypotheses. This demonstrates an ability to quickly simplify complex algebraic expressions and discharge goals using arithmetic reasoning.

Kimina-Prover's solution is also structured around proving the two conjuncts separately using $constructor$. It uses $have$ to establish intermediate inequalities and then applies tactics like $linarith$, $sq\_lt\_sq'$, $rw [Real.sq\_sqrt]$, and $nlinarith$. The steps are logical and clear but involve more explicit intermediate assertions compared to MPS-Prover.

DeepSeek-Prover V2's proof is the most detailed. It also uses $constructor$ (implicitly, by proving $h_3$ and $h_{11}$ separately for the conjuncts) and introduces many intermediate facts with $have$. The reasoning involves careful manipulation of inequalities, squaring both sides (e.g., $gcongr$), and algebraic simplification, often broken down into very small, verifiable steps. While demonstrating thoroughness, this level of granularity contributes to the proof's length.

This case again illustrates MPS-Prover's tendency towards conciseness by selecting powerful, progress-making tactics. The whole-proof methods, while effective, often produce more verbose proofs due to their strategy of explicit lemma introduction and detailed, step-by-step derivations.

\begin{minted}[fontsize=\small, breaklines, linenos, bgcolor=gray!10]{lean}
-- Solution to imo_1962_p2 by our MPS-Prover

import Mathlib
import Aesop
set_option maxHeartbeats 0
open BigOperators Real Nat Topology Rat

theorem imo_1962_p2 (x : ℝ) (h₀ : 0 ≤ 3 - x) (h₁ : 0 ≤ x + 1)
    (h₂ : 1 / 2 < Real.sqrt (3 - x) - Real.sqrt (x + 1)) : -1 ≤ x ∧ x < 1 - Real.sqrt 31 / 8 := by

  refine' ⟨_, _⟩
  nlinarith [Real.sq_sqrt h₀, Real.sq_sqrt h₁]
  rw [lt_sub_iff_add_lt] at h₂
  rw [lt_sqrt (by positivity)] at h₂
  rw [← sub_pos] at h₂ ⊢
  ring_nf
  ring_nf at h₀ h₁ h₂ ⊢
  field_simp at h₂ ⊢
<;> ring_nf at h₂ ⊢ <;> norm_num [h₀, h₁, h₂]
  rw [Real.sqrt_lt (by linarith)] <;> norm_num
<;> nlinarith [Real.sqrt_nonneg (1 + x), Real.sq_sqrt h₁]
\end{minted}

\begin{minted}[fontsize=\small, breaklines, linenos, bgcolor=gray!10]{lean}
-- Solution to imo_1962_p2 by Kimina-Prover

import Mathlib
import Aesop

set_option maxHeartbeats 0

open BigOperators Real Nat Topology Rat


theorem imo_1962_p2 (x : ℝ) (h₀ : 0 ≤ 3 - x) (h₁ : 0 ≤ x + 1)
    (h₂ : 1 / 2 < Real.sqrt (3 - x) - Real.sqrt (x + 1)) : -1 ≤ x ∧ x < 1 - Real.sqrt 31 / 8 := by 
  constructor
  · linarith [h₁]
  · have h3 : Real.sqrt (3 - x) - Real.sqrt (x + 1) > 1 / 2 := by linarith
    have h4 : Real.sqrt (3 - x) > Real.sqrt (x + 1) + 1 / 2 := by linarith
    have h5 : Real.sqrt (x + 1) ≥ 0 := Real.sqrt_nonneg (x + 1)
    have h6 : Real.sqrt (3 - x) ≥ 0 := Real.sqrt_nonneg (3 - x)
    have h7 : Real.sqrt (3 - x) > 0 := by linarith [h6, h5]
    have h8 : (Real.sqrt (3 - x)) ^ 2 > (Real.sqrt (x + 1) + 1 / 2) ^ 2 := by
      apply sq_lt_sq'
      all_goals linarith [h4, h5, h7]
    have h9 : (Real.sqrt (3 - x)) ^ 2 = 3 - x := by 
      rw [Real.sq_sqrt]
      linarith
    have h10 : (Real.sqrt (x + 1) + 1 / 2) ^ 2 = (Real.sqrt (x + 1)) ^ 2 + Real.sqrt (x + 1) + 1 / 4 := by
      ring_nf
      <;> simp [Real.sqrt_nonneg]
    rw [h9, h10] at h8
    have h11 : (Real.sqrt (x + 1)) ^ 2 = x + 1 := by
      rw [Real.sq_sqrt]
      linarith
    have h12 : Real.sqrt (x + 1) < 7 / 4 - 2 * x := by nlinarith [h8, h11, Real.sqrt_nonneg (x + 1)]
    have h13 : x < 1 - Real.sqrt 31 / 8 := by
      have h14 : Real.sqrt (x + 1) + 2 * x < 7 / 4 := by linarith [h12]
      have h15 : Real.sqrt 31 ^ 2 = 31 := by
        rw [Real.sq_sqrt]
        norm_num
      nlinarith [h14, Real.sq_sqrt (show 0 ≤ x + 1 by linarith), Real.sq_sqrt (show 0 ≤ 3 - x by linarith), Real.sqrt_nonneg (3 - x), Real.sqrt_nonneg (x + 1), Real.sq_sqrt (show 0 ≤ (31 : ℝ) by norm_num), Real.sq_sqrt (show 0 ≤ (8 : ℝ) by norm_num)]
    linarith [h13]
\end{minted}

\begin{minted}[fontsize=\small, breaklines, linenos, bgcolor=gray!10]{lean}
-- Solution to imo_1962_p2 by Deepseek-Prover V2

import Mathlib
import Aesop

set_option maxHeartbeats 0

open BigOperators Real Nat Topology Rat


theorem imo_1962_p2 (x : ℝ) (h₀ : 0 ≤ 3 - x) (h₁ : 0 ≤ x + 1)
    (h₂ : 1 / 2 < Real.sqrt (3 - x) - Real.sqrt (x + 1)) : -1 ≤ x ∧ x < 1 - Real.sqrt 31 / 8 := by
  have h₃ : -1 ≤ x := by
    linarith

  have h₄ : Real.sqrt (3 - x) - Real.sqrt (x + 1) > 1 / 2 := by
    linarith

  have h₅ : Real.sqrt (3 - x) > Real.sqrt (x + 1) + 1 / 2 := by
    linarith

  have h₆ : 3 - x > x + 1 + Real.sqrt (x + 1) + 1 / 4 := by
    have h₆₁ : Real.sqrt (3 - x) > Real.sqrt (x + 1) + 1 / 2 := h₅
    have h₆₂ : 0 ≤ Real.sqrt (x + 1) := Real.sqrt_nonneg (x + 1)
    have h₆₃ : 0 ≤ Real.sqrt (3 - x) := Real.sqrt_nonneg (3 - x)
    have h₆₄ : 0 ≤ Real.sqrt (x + 1) * (1 / 2) := by positivity
    have h₆₅ : 0 ≤ Real.sqrt (3 - x) := Real.sqrt_nonneg (3 - x)
    have h₆₆ : 0 ≤ Real.sqrt (x + 1) := Real.sqrt_nonneg (x + 1)
    have h₆₇ : (Real.sqrt (x + 1) + 1 / 2) ≥ 0 := by positivity
    have h₆₈ : Real.sqrt (3 - x) > Real.sqrt (x + 1) + 1 / 2 := h₅
    have h₆₉ : (Real.sqrt (x + 1) + 1 / 2) ^ 2 < (Real.sqrt (3 - x)) ^ 2 := by
      gcongr
      <;> nlinarith [Real.sq_sqrt (show 0 ≤ x + 1 by linarith), Real.sq_sqrt (show 0 ≤ 3 - x by linarith)]
    have h₆₁₀ : (Real.sqrt (x + 1) + 1 / 2) ^ 2 = (x + 1) + Real.sqrt (x + 1) + 1 / 4 := by
      nlinarith [Real.sq_sqrt (show 0 ≤ x + 1 by linarith), Real.sq_sqrt (show 0 ≤ 3 - x by linarith),
        sq_nonneg (Real.sqrt (x + 1) - 1 / 2)]
    have h₆₁₁ : (Real.sqrt (3 - x)) ^ 2 = 3 - x := by
      rw [Real.sq_sqrt] <;> linarith
    nlinarith [Real.sq_sqrt (show 0 ≤ x + 1 by linarith), Real.sq_sqrt (show 0 ≤ 3 - x by linarith)]

  have h₇ : 7 - 8 * x > 4 * Real.sqrt (x + 1) := by
    have h₇₁ : 3 - x > x + 1 + Real.sqrt (x + 1) + 1 / 4 := h₆
    have h₇₂ : 0 ≤ Real.sqrt (x + 1) := Real.sqrt_nonneg (x + 1)
    nlinarith [Real.sqrt_nonneg (x + 1)]

  have h₈ : x < 7 / 8 := by
    by_contra h
    have h₈₁ : x ≥ 7 / 8 := by linarith
    have h₈₂ : 7 - 8 * x ≤ 0 := by
      linarith
    have h₈₃ : 4 * Real.sqrt (x + 1) ≥ 0 := by positivity
    linarith

  have h₉ : 0 ≤ Real.sqrt 31 := by positivity

  have h₁₀ : 64 * x ^ 2 - 128 * x + 33 > 0 := by
    have h₁₀₁ : 7 - 8 * x > 4 * Real.sqrt (x + 1) := h₇
    have h₁₀₂ : 0 ≤ Real.sqrt (x + 1) := Real.sqrt_nonneg (x + 1)
    have h₁₀₃ : 0 ≤ 7 - 8 * x := by linarith
    have h₁₀₄ : 0 ≤ Real.sqrt (x + 1) := Real.sqrt_nonneg (x + 1)
    have h₁₀₅ : 0 ≤ 7 - 8 * x := by linarith
    have h₁₀₆ : (7 - 8 * x) ^ 2 > (4 * Real.sqrt (x + 1)) ^ 2 := by
      gcongr
    have h₁₀₇ : (4 * Real.sqrt (x + 1)) ^ 2 = 16 * (x + 1) := by
      nlinarith [Real.sq_sqrt (show 0 ≤ x + 1 by linarith), sq_nonneg (Real.sqrt (x + 1) - 2)]
    have h₁₀₈ : (7 - 8 * x) ^ 2 > 16 * (x + 1) := by
      linarith
    nlinarith [Real.sq_sqrt (show 0 ≤ x + 1 by linarith), sq_nonneg (x - 1)]

  have h₁₁ : x < 1 - Real.sqrt 31 / 8 := by
    by_contra h
    have h₁₁₁ : x ≥ 1 - Real.sqrt 31 / 8 := by
      linarith
    have h₁₁₂ : 64 * x ^ 2 - 128 * x + 33 ≤ 0 := by
      have h₁₁₃ : x ≥ 1 - Real.sqrt 31 / 8 := h₁₁₁
      have h₁₁₄ : x ≤ 1 + Real.sqrt 31 / 8 := by
        by_contra h₁₁₄
        have h₁₁₅ : x > 1 + Real.sqrt 31 / 8 := by linarith
        have h₁₁₆ : 64 * x ^ 2 - 128 * x + 33 > 0 := h₁₀
        have h₁₁₇ : x < 7 / 8 := h₈
        nlinarith [Real.sqrt_nonneg 31, Real.sq_sqrt (show 0 ≤ 31 by norm_num)]
      nlinarith [Real.sqrt_nonneg 31, Real.sq_sqrt (show 0 ≤ 31 by norm_num),
        sq_nonneg (x - (1 - Real.sqrt 31 / 8)), sq_nonneg (x - (1 + Real.sqrt 31 / 8))]
    linarith [h₁₀]

  exact ⟨h₃, h₁₁⟩
\end{minted}

\subsection{Analysis of a Problem Uniquely Solved by MPS-Prover: \texttt{amc12a\_2020\_p7}}

The problem \texttt{amc12a\_2020\_p7} involves evaluating a sum based on a sequence defined by specific values of $a(k)^3$. This problem was solved by our MPS-Prover, while both Kimina-Prover and DeepSeek-Prover V2 failed to find a solution.

The MPS-Prover solution employs $induction'$ as a key initial step. This is followed by extensive use of $simp\_all$ (often with $config := \{decide := true\}$) to simplify goals after instantiating terms of the sequence ($a(0)$ through $a(6)$) derived from the hypotheses $h_0$ through $h_6$. The proof also uses $interval\_cases$ to handle natural number variables within certain ranges derived from $nlinarith$. The final steps involve further simplification with $Nat.cast$ properties. The success here suggests that MPS-Prover's search was able to identify a productive high-level strategy ($induction'$) and then effectively use simplification and case analysis tactics to manage the resulting subgoals. The ability to find this particular combination of tactics, especially the crucial $induction'$ and effective use of $interval\_cases$, within the search budget highlights the strength of our multi-perspective approach in navigating complex search spaces where other methods might falter. The other provers might have struggled to identify the correct induction variable or effectively simplify the numerous concrete arithmetic subgoals that arise.

This case study highlights MPS-Prover's ability to find solutions to problems that prove difficult for other leading systems. Such successes can be attributed to the robust search paradigm of stepwise provers, which allows for exploration of various proof strategies, selection of effective tactics, and crucial backtracking capabilities. These features, amplified by our multi-perspective enhancements, enable the discovery of solutions even when the reasoning path is intricate or non-obvious.

\begin{minted}[fontsize=\small, breaklines, linenos, bgcolor=gray!10]{lean}
-- A problem solved by our MPS-Prover, where Kimina-Prover and Deepseek-Prover both fail to solve.

import Mathlib
import Aesop
set_option maxHeartbeats 0
open BigOperators Real Nat Topology Rat

theorem amc12a_2020_p7 (a : ℕ → ℕ) (h₀ : (a 0)^3 = 1) (h₁ : (a 1)^3 = 8) (h₂ : (a 2)^3 = 27) (h₃ : (a 3)^3 = 64) (h₄ : (a 4)^3 = 125) (h₅ : (a 5)^3 = 216) (h₆ : (a 6)^3 = 343) : ∑ k in Finset.range 7, (6 * (a k)^2) - ↑(2 * ∑ k in Finset.range 6, (a k)^2) = 658 := by 

  induction' 4 <;> simp_all [Finset.sum_range_succ, pow_succ]
  have h₇ : a 1 ≤ 8 := by nlinarith
  interval_cases a 1 <;> simp_all (config := {decide := true})
  have h₇ : a 2 * a 2 * a 2 = 27 := by assumption
  have h₈ : a 3 * a 3 * a 3 = 64 := h₃
<;> try simp_all [Nat.mul_comm, Nat.mul_assoc, Nat.mul_left_comm]
  have h₉ : a 4 * a 4 * a 4 = 125 := by nlinarith
  all_goals
  have : a 2 ≤ 6 := by nlinarith
  interval_cases a 2 <;> simp_all (config := {decide := true})
  have h₁₀ : a 5 * a 5 * a 5 = 216 := by nlinarith
  have h₁₂ : a 6 = 7 := by nlinarith
  simp_all
  all_goals
  have : a 4 ≤ 12 := by nlinarith
  interval_cases a 4 <;> simp_all (config := {decide := true})
  have h₁₄ : a 5 = 6 := by nlinarith
  simp_all [Nat.cast_add, Nat.cast_mul, Nat.cast_pow]
  have h₁₁ : a 3 ≤ 8 := by nlinarith
  interval_cases a 3 <;> simp_all
\end{minted}

\end{document}